\documentclass[conference]{IEEEtran}
\IEEEoverridecommandlockouts

\usepackage{fancyhdr} 
\usepackage{booktabs}
\usepackage{multirow}
\usepackage{subfigure}
\usepackage{cite}
\usepackage{url}
\usepackage{amsmath,amssymb,amsfonts}
\usepackage{algorithmic}
\usepackage{graphicx}
\usepackage{textcomp}
\usepackage{xcolor}
\def\BibTeX{{\rm B\kern-.05em{\sc i\kern-.025em b}\kern-.08em
    T\kern-.1667em\lower.7ex\hbox{E}\kern-.125emX}}
\begin{document}

\newcommand{\major}[1]{\textcolor{black}{#1}}

\title{Understanding Time Variations of DNN Inference in Autonomous Driving}

\author{
\IEEEauthorblockN{
Liangkai~Liu\IEEEauthorrefmark{1},
Yanzhi~Wang\IEEEauthorrefmark{2}, and
Weisong~Shi\IEEEauthorrefmark{3}
}
\IEEEauthorblockA{
\IEEEauthorrefmark{1}Department of Computer Science, Wayne State University
\\
\IEEEauthorrefmark{2}Department of Electrical \& Computer Engineering, Northeastern University
\\
\IEEEauthorrefmark{3}Department of Computer and Information Sciences, University of Delaware\\
}
}

\maketitle

\thispagestyle{fancy} 
\lhead{} 
\chead{} 
\rhead{} 
\lfoot{} 
\cfoot{\thepage} 
\rfoot{} 
\renewcommand{\headrulewidth}{0pt} 
\renewcommand{\footrulewidth}{0pt} 

\pagestyle{fancy}  
\cfoot{\thepage}

\begin{abstract}
  Deep neural networks (DNNs) are widely used in autonomous driving due to their high accuracy for perception, decision, and control. In safety-critical systems like autonomous driving, executing tasks like sensing and perception in real-time is vital to the vehicle’s safety, which requires the application’s execution time to be predictable. However, non-negligible time variations are observed in DNN inference. Current DNN inference studies either ignore the time variation issue or rely on the scheduler to handle it. None of the current work explains the root causes of DNN inference time variations. Understanding the time variations of the DNN inference becomes a fundamental challenge in real-time scheduling for autonomous driving. \major{In this work, we analyze the time variation in DNN inference in fine granularity from six perspectives: data, I/O, model, runtime, hardware, and end-to-end perception system. Six insights are derived in understanding the time variations for DNN inference.} 
\end{abstract}

\begin{IEEEkeywords}
time variations, deep neural networks, autonomous driving
\end{IEEEkeywords}

\section{Introduction}
\label{introduction}

Owing to its high safety and efficiency, autonomous driving has become the fundamental technology for the next generation of transportation. Deep Neural Networks (DNNs) are widely deployed in the autonomous driving system for sensing, perception, decision, and control. Typical examples include: YOLOv3 and Faster R-CNN for object detection~\cite{redmon2018yolov3,ren2015faster}; Deeplabv3 for image semantic segmentation~\cite{howard2017mobilenets,chen2017rethinking}; LaneNet and PINet for lane detection~\cite{neven2018towards, ko2020key}. There are two main reasons for the success of DNNs in autonomous driving systems. The first is the higher accuracy compared with traditional computer vision-based approaches~\cite{grigorescu2020survey}. The other is that DNNs can process raw data, making it suitable for autonomous driving vehicles since it generates terabytes of raw sensor data daily~\cite{liu2020computing}. 

As a safety-critical system, autonomous driving sets high requirements in accuracy, real-time, robustness, etc. The high accuracy of DNN-based algorithms promotes the development of autonomous vehicles. However, satisfying the real-time requirements of the sensing, perception, and planning tasks are still significant challenges. According to~\cite{kato2015open}, when the vehicle drives at 40 km per hour in urban areas, these autonomous functions should be effective every 1 m with task execution time less than 100ms. As DNN models are widely used in object detection/classification, lane tracking, and decision-making applications, guaranteeing the real-time execution of the DNN inference becomes the key to satisfying the real-time requirements of autonomous driving.

Generally, for safety-critical applications like sensing, perception, control, etc., deadline-based scheduling is used by the real-time scheduler to guarantee safety. Setting up deadlines is usually based on the worst observed execution time. However, although the model structure and weights are fixed, non-negligible time variations still exist in DNN inference~\cite{wu2019machine,wan2020alert}. Prior works~\cite{wu2019machine} observed the time variations for DNN inference in mobile devices and found that inference time follows an approximately Gaussian distribution. However, the statistic-based approach performs poorly when time variations are enormous. Another work~\cite{wan2020alert} on the anytime DNN system also observed the time variations issue and provided a Kalman Filter-based estimation for latency distribution. $D^3$ is a work that addresses the time variations in AV systems with dynamic deadlines rather than static deadlines~\cite{gog2022d3}. However, none of the existing approaches could handle huge time variations since they do not consider the roots causing DNN inference time variations. 
Time variations bring a big challenge for real-time schedulers because the deadline with the worst observed execution time could waste many processor resources. Many resources would be saved if the DNN inference time could be fixed or narrowed down to a specific range. A detailed and in-depth analysis of DNN inference time variation is missing. Therefore, \textit{understanding the DNN inference time variation is key to optimizing the DNN inference runtime for autonomous driving}. 

In this paper, we undertake a comprehensive analysis of DNN inference time variations in a general autonomous driving system. We analyze the time variation issues in fine granularity. For typical DNN models, we consider the variability in DNN inference from six perspectives: \textit{data, I/O, model, runtime,} \textit{hardware}, and end-to-end perception system. Six insights are derived for reducing DNN inference time variations. 
In summary, this paper makes the following contributions: 
\begin{itemize}
    \item The time variation issues of DNN inference in autonomous driving are thoroughly studied. We found that the majority of DNN models show variations larger than 100ms, which significantly affects autonomous driving safety.
    \item Through a comprehensive analysis of the time variation of DNN model inference from six perspectives, we derive six insights into the relationship between DNN inference time variations and variability of data, I/O, model, runtime, hardware, and end-to-end perception system.
\end{itemize}

The rest of the paper is organized as follows. Section~\ref{motivation} presents the background and motivation of this work. Section~\ref{profiling} discusses the time variation analysis for typical DNN models. Section~\ref{implementation} present the implementation of the end-to-end perception system. Section~\ref{system-analysis} dicussed the time variation analysis for the end-to-end perception system. 
Section~\ref{related-work} describes the related work. Section~\ref{Conclusion} concludes the paper. 

\section{Background and Motivation}
\label{motivation}

The recent proliferation of computing technologies, e.g., sensors, computer vision, machine learning, hardware acceleration, and the broad deployment of communication mechanisms, e.g., DSRC, C-V2X, 5G, have pushed the horizon of autonomous driving, which automates the decision and control of vehicles by leveraging the perception results based on multiple sensors. The key to the success of these autonomous systems is making reliable decisions in a real-time fashion. However, since the sensing data and runtime of the computing devices vary, the end-to-end latency for the perception, planning, and vehicle controls show non-negligible variation. This time variation affects the system predictability, making the operating system's scheduling less efficient. 

In this section, we present the motivation for this work by answering three questions:

\begin{itemize}
    \item How prevalent are DNNs for autonomous driving?
    \item How vital is timing variation for autonomous driving, and what is the state-of-the-art?
    \item What are the potential issues that affected the time variations in DNN inference?
\end{itemize}

\subsection{DNN Inference for Autonomous Driving}

\begin{figure}[!t]
	\centering
	\includegraphics[width=.8\columnwidth]{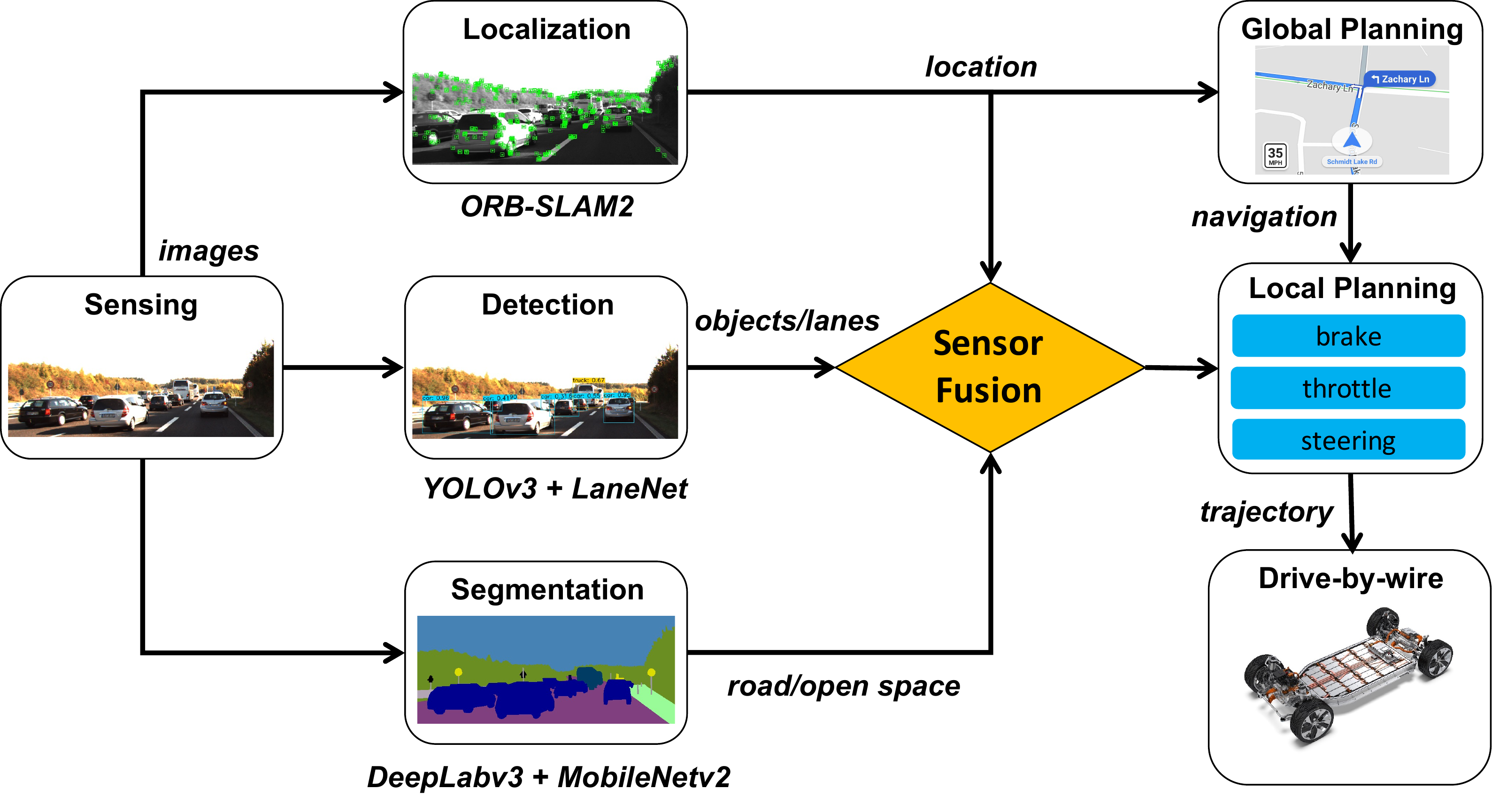}
	\caption{A typical autonomous driving system.}
	\label{fig:AD-system-overview}
\end{figure}

DNN models have been widely used in the autonomous driving system for sensing, perception, and localization~\cite{liu2019edge}. Figure~\ref{fig:AD-system-overview} shows an overview of the state-of-the-art autonomous driving systems. The design purely relies on cameras for sensing the environment. As shown in the figure, eight primary components are divided into three parts: sensing, perception, and decision~\cite{liu2017creating}. Sensing is the process that sensors capture information from the environment. Perception represents understanding the environment with algorithms applied to the sensing data, including localization, detection, semantic segmentation, and sensor fusion. The role of sensor fusion is to collaborate among all the perception components and generate locations for objects, lanes, and open spaces for the planning module. The decision is composed of global planning, local planning, and vehicle control. Global planning generates the routes between origin and destination, while local planning generates the trajectory and control commands on brake, throttle, and steering. Finally, the control commands are sent to the drive-by-wire system and applied to the vehicle. 

We can find that the core modules on perception and decision are all based on neural networks. The main reason is that DNN has an excellent performance in many applications. Take image classification as an example. The ImageNet challenge winner in 2017 decreased the classification error to 0.023, while human beings' error is 0.05~\cite{krizhevsky2012imagenet}. DNNs can also learn features from raw, heterogeneous, and noisy data, suitable for autonomous driving scenarios since terabytes of raw sensor data are generated on the vehicle~\cite{dean20201}.

\subsection{Time Variations in DNN Inference}

The proliferation of deep learning achieves enormous performance improvement and brings lots of issues in computation complexity, energy consumption, and time variation for safety-critical systems~\cite{wu2019machine}. For autonomous driving vehicles, time variation affects predictability, which guarantees the vehicle's safety and improves resource utilization~\cite{park1992predicting}.


However, current DNN-based computing systems show poor performance in terms of time variation~\cite{liu2019e2m}. To illustrate the time variation issue in a state-of-the-art DNN-based autonomous driving system, we choose eleven models/algorithms covering the whole pipeline of the autonomous driving system and measure the end-to-end latency with the same input. There are seven DNN models for perception: YOLOv3~\cite{redmon2018yolov3}, Faster R-CNN ResNet101~\cite{ren2015faster}, Mask R-CNN Inceptionv2~\cite{he2017mask}, and SSD MobileNetv2~\cite{liu2016ssd,howard2017mobilenets} are for object detection; PINet~\cite{ko2020key} and LaneNet~\cite{neven2018towards} for lane detection; Deeplabv3 with MobileNetv2 for semantic segmentation~\cite{chen2017rethinking,howard2017mobilenets}. Localization and planning algorithms are tested by running ROS Navigation offline with recorded sensor data~\cite{ros-navigation}. Adaptive Monte Carlo localization (AMCL)~\cite{ros-amcl} and ORB-SLAM2~\cite{mur2016orb} are deployed for localization. A$^*$~\cite{ros-global-planner} and Dynamic Window Approach (DWA)~\cite{ros-local-planner} are deployed for global and local path planning, respectively. Table~\ref{tab:latency-motivation} shows the mean, the range, and the division of the range and the mean of the end-to-end latency for eleven models/algorithms. The range is defined as the difference between the maximum value and the minimum value. From Table~\ref{tab:latency-motivation}, we can observe that majority of the time is consumed by the DNN-based perception tasks: object detection, lane detection, and segmentation. Among all the seven DNN models, four models have a range larger than 100ms. LaneNet shows the biggest range with 282ms. If we consider the percentage of the range over the mean, the variations of the last three models for lane detection, localization, and planning are larger than 100 percent. AMCL shows the poorest performance with 675.5 percent. However, localization and planning tasks only occupy a limited portion of the end-to-end autonomous driving pipeline. 

\begin{table}
\centering
\caption{The mean, range, and variations of the eleven models used in the autonomous driving system.}
\label{tab:latency-motivation}
\resizebox{\linewidth}{!}{
\begin{tabular}{ccccc} 
\toprule
\textbf{Task} & \textbf{Model} & \textbf{Mean (ms)} & \textbf{Range (ms)} & \textbf{Range / Mean (\%)} \\ 
\midrule
\multirow{4}{*}{Object Detection} & YOLOv3~\cite{redmon2018yolov3} & \textbf{173} & 57 & 32.8 \\
 & Faster R-CNN Resnet101~\cite{ren2015faster} & \textbf{413} & \textbf{128} & 31 \\
 & Mask R-CNN Inceptionv2~\cite{he2017mask} & \textbf{266} & \textbf{104} & 39.1 \\
 & SSD MobileNetv2~\cite{liu2016ssd,howard2017mobilenets} & \textbf{144} & 70 & 48.6 \\ 
\midrule
\multirow{2}{*}{Lane Detection} & LaneNet~\cite{neven2018towards} & 82 & \textbf{282 } & \textbf{344} \\
 & PINet~\cite{ko2020key} & \textbf{127} & \textbf{263} & \textbf{207.1} \\ 
\midrule
Segmentation & Deeplabv3~\cite{chen2017rethinking,howard2017mobilenets} & \textbf{149} & 19 & 12.8 \\ 
\midrule
\multirow{2}{*}{Localization} & AMCL~\cite{ros-amcl} & 1.3  & 8.7 & \textbf{675.5} \\
 & ORB-SLAM2~\cite{mur2016orb} & 53 & 56 & \textbf{105.6} \\ 
\midrule
\multirow{2}{*}{Planning} & A*~\cite{ros-global-planner} & 79 & 97 & \textbf{122.3} \\
 & DWA~\cite{ros-local-planner} & 23 & 73 & \textbf{323.8} \\ 
\bottomrule
\end{tabular}
}
\end{table}




\begin{figure}[!htp]
	\centering
	\includegraphics[width=\columnwidth]{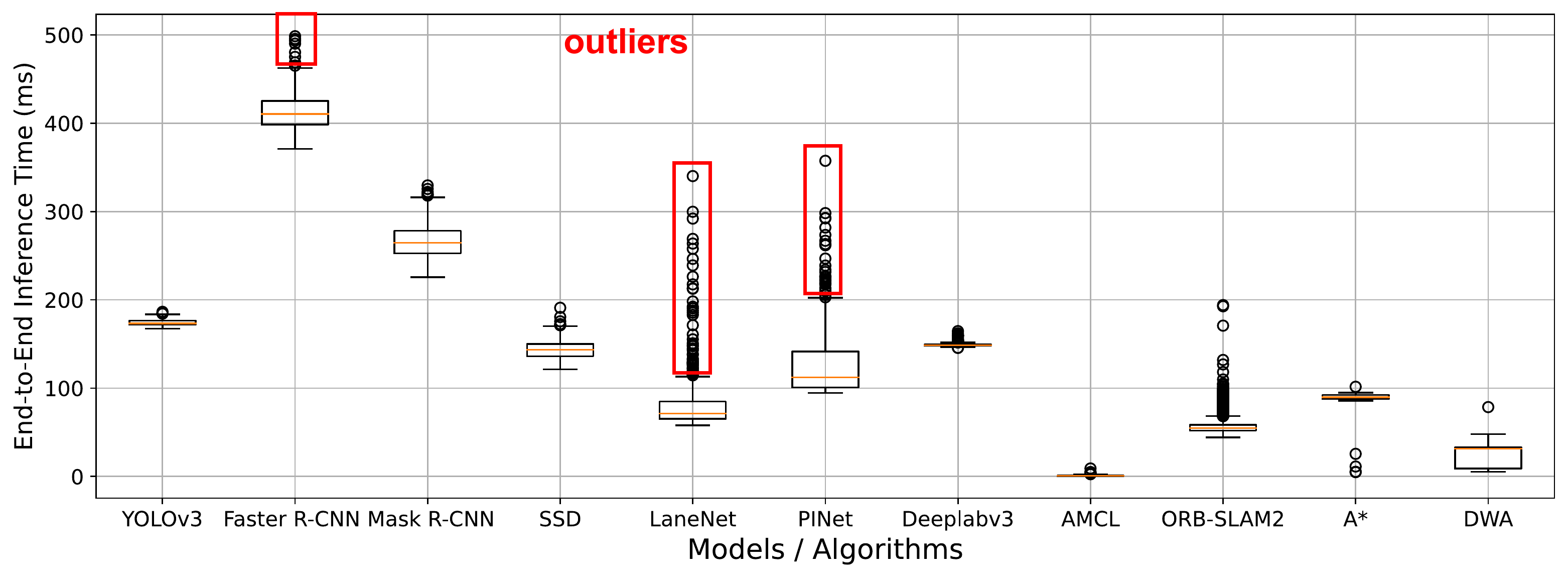}
	\caption{The box plot of end-to-end latency for typical models/algorithms in autonomous driving pipeline.}
	\label{fig:e2e-boxplot}
\end{figure}

How about the distributions of the end-to-end latency? Figure~\ref{fig:e2e-boxplot} shows the box plot of the end-to-end inference time of the above eleven models/algorithms. We can observe that perception tasks have a much wider latency distribution than localization and planning tasks. Non-negligible amount of latencies that lie above the 75th percentile. Besides, there are many outliers (abnormal data) among all the perception tasks. Although these outliers have a low probability, they affect the scheduler's performance, especially for the safety-critical system like autonomous driving. If we consider the autonomous driving system a hard real-time system, the scheduler assigns deadlines for each task based on the worst-observed execution time. Take LaneNet as an example. The scheduler would set the deadline larger than the worst observed execution time, which is 340ms~\cite{wilhelm2008worst}. Although it guarantees the safety of this task, it also brings enormous inefficiency because the actual execution time is less than 160ms over 95 percent of the time, which means around 180ms is wasted for most jobs. If we narrow the execution time range to less than 50ms, we can save almost 110ms in LaneNet for every job.

Time variation is expected to make a significant impact on safety-critical systems. Since perception tasks consume most of the time, this work's focus would be on the DNN models in perception of the autonomous driving system. We propose using profiling tools to understand DNN inference time variations in current autonomous driving systems and model their time variation.

\subsection{Uncertainties in DNN Inference}

\begin{figure}[!htp]
	\centering
	\includegraphics[width=\columnwidth]{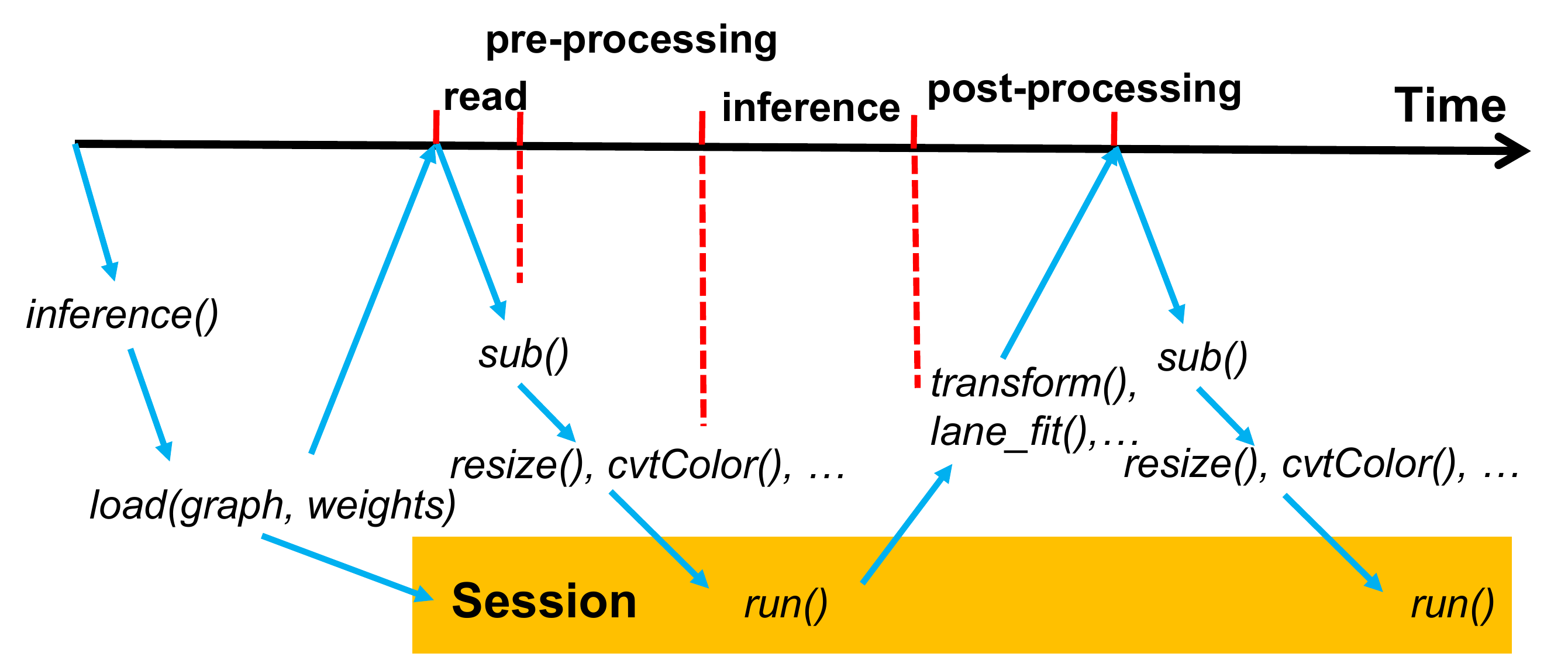}
	\caption{The timeline of DNN inference.}
	\label{fig:dnn-timeline}
\end{figure}

What are the potential issues that affect the time variations in DNN inference? To answer this question, we first analyze the timeline of DNN inference. Figure~\ref{fig:dnn-timeline} shows a general DNN inference timeline in TensorFlow. The timeline starts by calling \textit{inference()}, which first loads the graphs and weights into the memory. The process starts reading the input, where \textit{sub()} is used to subscribe to an image stream. The Robot Operating System~\cite{quigley2009ros} (ROS) provides data communications between different components. Next, it pre-processes the image, including resizing, converting color space from one to another, etc. Next, the processed image is passed to the \textit{Session} and loaded into the processor to run network inference. Finally, post-processing transforms the bounding boxes \textit{transform()} or fit pixel proposals into lanes (\textit{lane\_fit()}) on the image.

From the timeline of DNN inference, we derive several uncertainties that contribute to the time variation issue. The first is data, which means the value and distribution of image pixels. The sparsity matrix is expected to have less time on inference than the dense matrix~\cite{chen2016eyeriss}. The second is the data I/O to the session graph. How the running graph reads data from ROS messages could affect the inference time variations. The third is the model, which means the model's structure and complexity in multiply and accumulate (MAC) operations. The fourth is runtime, owing to the contention of concurrent jobs for resources like memory, CPU, and GPU. How many processes allow preemption, their scheduling policies, priorities, etc., are the factors that affect the runtime variation. Finally, the hardware also affects the time variation of DNN inference. GPU is expected to run faster than CPU, but a multi-core system is supposed to show more time variations than a single-core system. How will different architectures affect the time variations of DNN inference? In summary, five aspects of uncertainties in DNN inference need to be studied: data, I/O, model, runtime, and hardware.

\subsection{Profiling Tools}

For current DNN-based autonomous driving systems, a big challenge is how to explore the time variation in DNN model inference. Our approach is to profile the system with a variety of granularity. In general, we use three profiling tools: code level (\textit{cProfiler}) and system level (\textit{nvprof} and \textit{Linux Perf}).



\textit{cProfiler}~\cite{python-profiler} is a python library for code-level profiling. It collects statistics that describe how often and for how long parts of the program are executed. The number of function calls can identify code bugs and possible inline-expansion points (high call counts). Internal time statistics can identify “hot loops” that should be carefully optimized. Cumulative time statistics are used to identify high-level errors in the selection of algorithms. We use \textit{cProfiler} to get the call graph with time breakdowns of the code.

Although application-level profiling gives us some explanations of DNN inference time variations, it is not enough to explain the variability of the DNN inference for different architectures. Therefore, we conduct system call level profiling with \textit{Linux perf} to show model inference performance at the system call level. \textit{Linux perf} uses the system performance counter to monitor the whole system \cite{gregg2013linux}. We use \textit{Linux perf} to collect the system-level metrics, including CPU cycles, context switches, CPU migrations, page faults, instructions, branches, branch misses, cache misses, etc.

GPU has been widely used in DNN execution acceleration. \textit{nvprof} is a profiling tool provided by NVIDIA that enables the collection of a timeline of CUDA-related activities on CPU and GPU, including kernel execution, memory transfers, memory set, CUDA API calls, and events or metrics for CUDA kernels~\cite{bradley2012gpu}. We use \textit{nvprof} to collect timing analysis for GPU activities to find the roots for DNN inference time variations.

\section{Model Inference Profiling}
\label{profiling}

To understand the uncertainties and quantify their impacts on DNN inference time variations, we need to profile the execution of model inference in fine granularity. This section discusses the profiling of typical DNN models based on the uncertainty analysis in Section~\ref{motivation}. 

\subsection{Experiment Setup}

To begin with, we present the experimental setup for DNN inference profiling. We choose three types of computing devices that cover CPU and GPU-based processors. Besides, we create an image dataset based on the KITTI dataset~\cite{geiger2012we}.

\vspace{0.3em}
\noindent\textbf{Hardware and software setup.} The devices we use for profiling include NVIDIA Jetson AGX Xavier, Xavier NX, and Intel Fog Reference. Table~\ref{tab:hardware-description} shows these devices' memory, CPU, and GPU configurations~\cite{jetson-comparison}. Both the Jetson boards are installed with \texttt{JetPack 4.4-DP}~\cite{nvidia-jetpack} (L4T R32.4.2) and use an \texttt{l4t-ml} docker image~\cite{l4tml-docker} as the base image for system setup. The \texttt{l4t-ml} image includes several libraries for machine learning-related applications: \texttt{TensorFlow 1.15}, \texttt{PyTorch v1.5.0}, \texttt{torchvision v0.6.0}, \texttt{CUDA 10.2}, \texttt{cuDNN 8.0.0}, \texttt{OpenCV 4.1}, etc. On top of the \texttt{l4t-ml} image, we implemented a ROS-based perception pipeline for autonomous driving. \texttt{ROS Melodic} and \texttt{ROS Galactic} are deployed as the communication middleware. To reduce the impact of different hardware on the time variation, the model inference profiling of data, I/O, model, and runtime variances are conducted on NVIDIA Jetson AGX boards, which have the same chip as the auto-graded NVIDIA DRIVE AGX Xavier board~\cite{nvidia-drive}. Besides, we disable the Dynamic Voltage and Frequency Scaling (DVFS) on Jetson AGX Xavier with \texttt{jetson\_benchmark}~\cite{jetson-benchmark} and turn off all other user applications before the experiment. The Jetson broad is set at \texttt{MAXN} power mode, where both CPU and GPU run with the highest frequency. Since accuracy is essential for the autonomous driving scenario, all the DNN models are trained and tested with full precision (FP32).

\begin{table*}
\centering
\caption{\major{Hardware configurations of devices in profiling.}}
\label{tab:hardware-description}
\footnotesize
\resizebox{\linewidth}{!}{%
\begin{tabular}{ccccc} 
\toprule
\textbf{Devices} & \textbf{CPU} & \textbf{GPU} & \textbf{Memory} & \textbf{AI Performance} \\ 
\hline
AGX Xavier & \begin{tabular}[c]{@{}c@{}}8-core NVIDIA Carmel Arm®v8.2 \\64-bit CPU 8MB L2 + 4MB L3\end{tabular} & \begin{tabular}[c]{@{}c@{}}512-core NVIDIA Volta™\\~with 64 Tensor Cores\end{tabular} & \begin{tabular}[c]{@{}c@{}}32 GB 256-bit LPDDR4x\\136.5 GB/s\end{tabular} & 32 TOPs \\ 
\hline
Xavier NX & \begin{tabular}[c]{@{}c@{}}6-core NVIDIA Carmel ARM®v8.2 \\64-bit CPU 6MB L2 + 4MB L3\end{tabular} & \begin{tabular}[c]{@{}c@{}}384-core NVIDIA Volta™\\~with 48 Tensor Cores\end{tabular} & \begin{tabular}[c]{@{}c@{}}8 GB 128-bit LPDDR4x\\59.7 GB/s\end{tabular} & 21 TOPs \\ 
\hline
Fog Node & \begin{tabular}[c]{@{}c@{}}8 Intel(R) Xeon(R) \\CPU E3-1275 v5 @ 3.60GHz\end{tabular} & - & \begin{tabular}[c]{@{}c@{}}32 GB DDR4\\34.1 GB/s\end{tabular} & - \\
\hline
GPU Workstation & \begin{tabular}[c]{@{}c@{}} 28 Intel® Core™ i9-9940X \\CPU @ 3.30GHz\end{tabular} & 4 NVIDIA GeForce RTX 2080 Ti/PCIe/SSE2 & \begin{tabular}[c]{@{}c@{}}64 GB DDR4\\85 GB/s \end{tabular} & 312 TOPS \\
\bottomrule
\end{tabular}
}
\end{table*}


\vspace{0.3em}
\noindent\textbf{Dataset Descriptions.} We create an image dataset as a uniform input to the profiling process based on the KITTI dataset~\cite{kitti-benchmark}. The image dataset covers three scenarios (city, residential, and road) with 1,800 images. By sampling the image dataset with different frequencies (1, 2, 5, and 10 FPS), we get four groups of images for each scenario, with 60, 120, 300, and 600 images for each scenario. In addition to the image dataset, an auto-grade camera with 1920x1080 resolution and 30 FPS is also used as input to the DNNs. 

\vspace{0.3em}
\noindent\textbf{Metrics.} To give detailed profiling results of the DNN-based applications, we measure several metrics, including latency, processor utilization, memory utilization, etc. We calculate latency's statistic metrics, including the range and the coefficient of variation. Equation~(\ref{equation:range}) defines the range as the difference between the maximum and minimum values. The coefficient of variation ($c_{v}$) is used to evaluate the relative variability, and it is calculated using the standard variation $\sigma$ divided by the mean value $\mu$. $c_{v}$ is a positive value. The higher the value is, the higher variations the data has. 

\textbf{Range:}
\begin{equation}
\label{equation:range}
R = max(t_{i}) - min(t_{i})
\end{equation}


\textbf{Coefficient of Variation:}
\begin{equation}
\label{equation:cv}
c_{v} = \frac{\sigma }{\mu }
\end{equation}



\subsection{Data Variability}






As the input to the DNN inference, data variability is expected to affect the whole pipeline significantly. The application scenarios and FPS are expected to affect the execution time of the DNN models since more lanes and objects are supposed to be detected downtown than in the countryside. Meanwhile, since the critical operations of DNN inference are multiply and accumulate (MAC), the distribution of the matrix/pixel might also affect the execution time. One example is that the sparse matrix needs fewer operations than the dense matrix. Moreover, the weather brings another uncertainty for sensor data when the model is deployed in real-world environment. Therefore, we discuss the impact of the data variability on the time variation of DNN inference in three aspects: scenarios, pixel sizes \& distribution, and weather's impact.

\begin{figure}[!htp]
	\centering
	\includegraphics[width=\columnwidth]{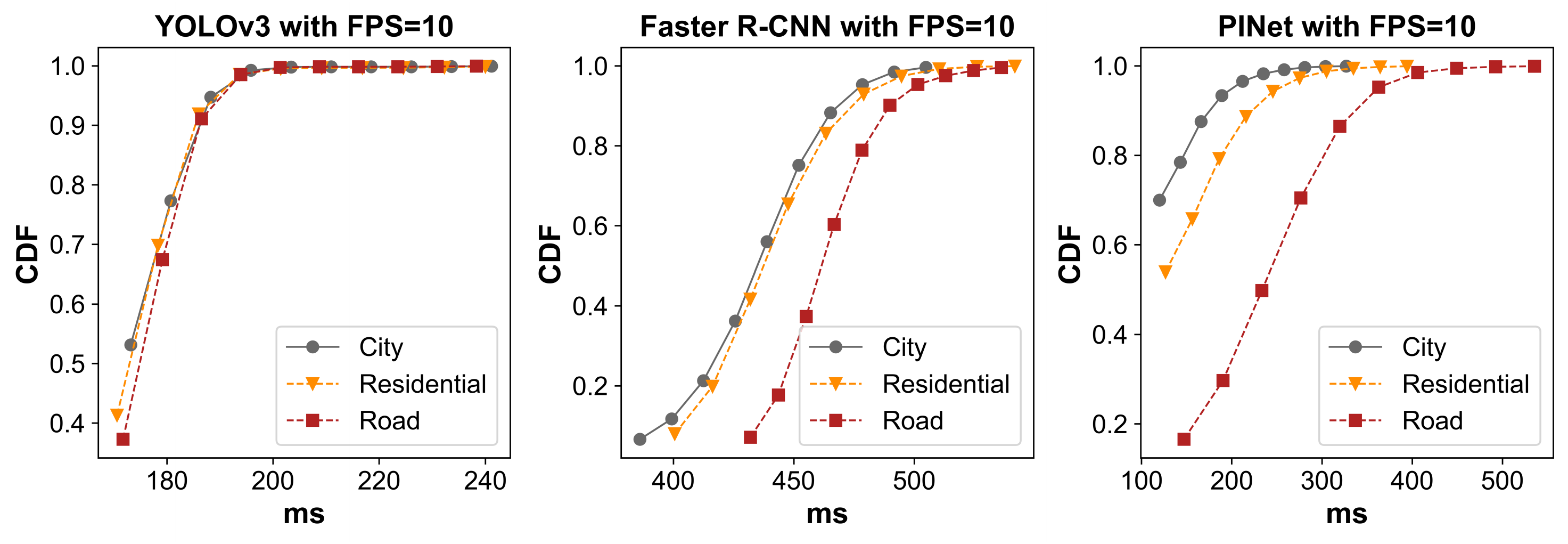}
	 \caption{CDF for latency results of YOLOv3, Faster R-CNN, and PINet under different scenarios.}
    \label{fig:data-scenario}
\end{figure}

\vspace{0.3em}
\major{\noindent\textbf{Scenarios.} We use the datasets from three scenarios (i.e., city, residential, and road) to evaluate the impact of data variability. Three DNN inferences are covered: YOLOv3 and Faster R-CNN for object detection, PINet for lane detection. The results under different scenarios are shown in Figure~\ref{fig:data-scenario}.} 
The performance for different scenarios shows differently: PINet and Faster R-CNN show massive time variations between those three scenarios, while YOLOv3 does not. The reason is that different scenarios bring variable possibilities to detect lanes and objects. As a representative of one-stage based object detection, YOLOv3's latency is not affected by the scenarios. However, Faster R-CNN is a two-stage based object detection. Different scenarios have different numbers of potential objects, contributing to the inference time variations. This observation also implies that the scheduling of object and lane detection tasks should take the running scenario into account. We will discuss it in detail in model variability. 



\begin{figure}[!htp]
	\centering
	\includegraphics[width=\columnwidth]{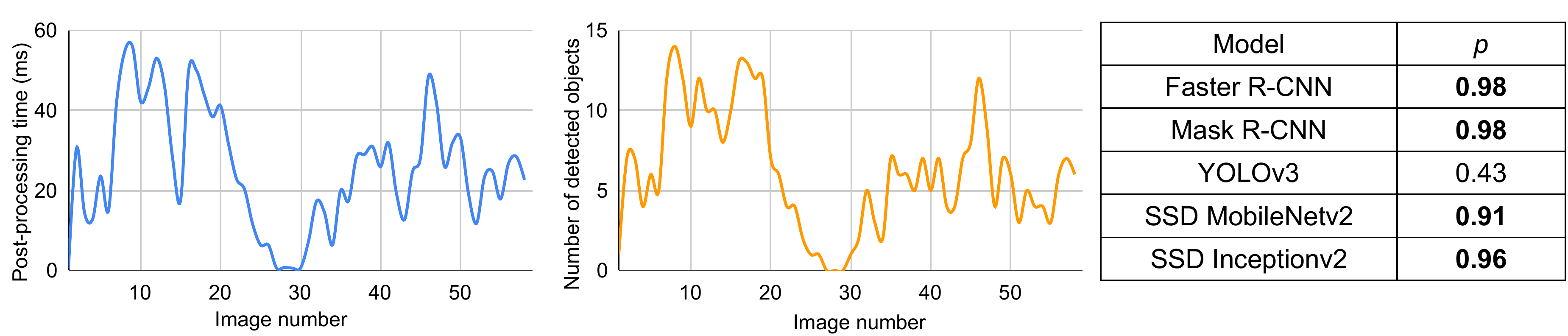}
	 \caption{The post-processing time and the number of detected objects in Faster R-CNN ResNet101 for a sequence of images. Right table contains the correlation coefficient of post-processing time and number of detected objects for five object detection models.}
    \label{fig:data-obj-post}
\end{figure}

Since the different scenarios mainly affect the possibility of detecting objects, the number of objects becomes a potential connection between the scenario and the time variation of DNN inference. To prove it, we choose several DNN models for object detection and get the time sequence of the latency breakdown and the number of objects. Figure~\ref{fig:data-obj-post} shows an example of the results for Faster R-CNN ResNet101. The variation in the number of objects is almost consistent with the variations in the post-processing time. To get more accurate correlation results, we calculate the correlation coefficient ($p$) of five object detection models' post-processing and the number of detected objects. The results for Faster R-CNN, Mask R-CNN, YOLOv3, SSD MobileNetv2, and SSD Inceptionv2 are 0.98, 0.98, 0.43, 0.91, and 0.96, respectively. These results explain the relationship between usage scenarios and inference time variations. Since lane detection can be seen as pixel-level regression, the scenarios' impact on its DNN inference is similar.

\begin{figure}[!htp]
	\centering
	\includegraphics[width=.9\columnwidth]{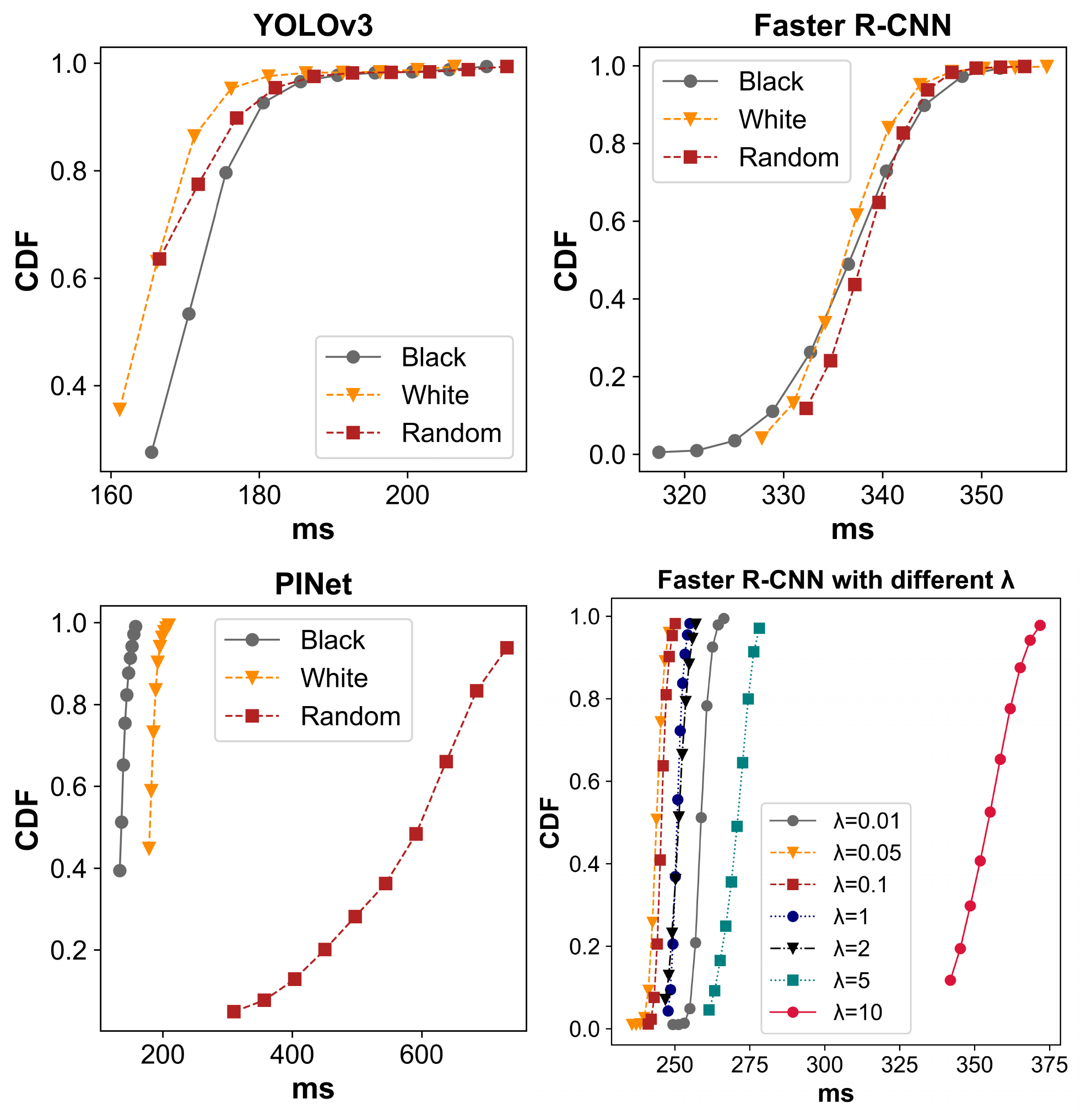}
	\caption{The CDF of end-to-end latency for YOLOv3, Faster R-CNN, and PINet with different pixel distributions; The CDF of end-to-end latency for Faster R-CNN with difference image sizes.}
	\label{fig:data-distribution}
\end{figure}

\vspace{0.3em}
\major{\noindent\textbf{Pixel sizes \& distributions.}} In addition to the scenarios of the input data, the pixel distribution and sizes are also expected to affect the variation of DNN inference time. To show the effect of pixel distributions, we choose three types of data inputs to the DNNs and measure time variations. These image inputs include all zero (black), all 255 (white), and random matrix. We get the results for running these three cases on YOLOv3, Faster R-CNN, and PINet, as shown in Figure~\ref{fig:data-distribution}. We can observe significant time variation in PINet when the random matrix is applied because PINet is a pixel-level regression to lanes. Random values of the pixels add the computations for detecting pixel proposals for lanes. On the contrary, there is no significant difference between YOLOv3 and Faster R-CNN because object detection is a box-level detection. Changing limited pixels does not make a big difference for a box containing thousands of pixels. 

\major{For profiling pixel size's impact on inference time variations, we choose the Faster R-CNN model and scale the same input image ([3, 375, 1242]) by multiplying the width and heights with different ratios $\lambda$: 0.01, 0.05, 0.1, 1, 2, 5, and 10. Each scaled image is sent to execute model inference 100 times. The results for CDF of the end-to-end latency are shown in Figure~\ref{fig:data-distribution}. We can find that the CDFs are close to each other except for the case when $\lambda$ equals 10, which shows higher average latency and inference time variations than others. Through the study on the implementation, we found that the difference is mainly caused by the pre-processing, where the input image will be transposed and cropped if the size is larger than the maximum value.}

\begin{table}
\centering
\refstepcounter{table}
\caption{The mean ($\mu$), variation ($\sigma$), and variation coefficient ($c_{v}$) of the end-to-end inference time for Faster R-CNN and PINet under different raining cases.}
\resizebox{\linewidth}{!}{%
\label{tab:rain-e2e}
\begin{tabular}{ccccccc} 
\toprule
End-to-End Inference Time (ms) & \multicolumn{3}{c|}{\textbf{Faster R-CNN}} & \multicolumn{3}{c}{\textbf{PINet}} \\ 
\hline
Raining Case & $\mu$ & $\sigma$ & \multicolumn{1}{c|}{$c_{v}$} & $\mu$ & $\sigma$ & $c_{v}$ \\ 
\hline
\textit{0 mm/hour} & 320.7 & 1217.9 & 3.8 & 228.9 & 7862.6 & 34.4 \\
\textit{25 mm/hour} & 326.5 & 1129.6 & 3.5 & 216.9 & 7142.3 & 32.9 \\
\textit{50 mm/hour} & 324.3 & 1082.5 & 3.3 & 215 & 7033.4 & 32.7 \\
\textit{100 mm/hour} & 319.3 & 950.4 & 3 & 205.9 & 6633.3 & 32.2 \\
\textit{150 mm/hour} & 314.9 & 898.9 & 2.9 & 199.2 & 5410.6 & 27.2 \\
\textit{200 mm/hour} & \textbf{309.1} & \textbf{805.5} & \textbf{2.6} & \textbf{190.4} & \textbf{5068.1} & \textbf{26.6} \\
\bottomrule
\end{tabular}
}
\end{table}

\begin{figure}[!htp]
	\centering
	\includegraphics[width=.9\columnwidth]{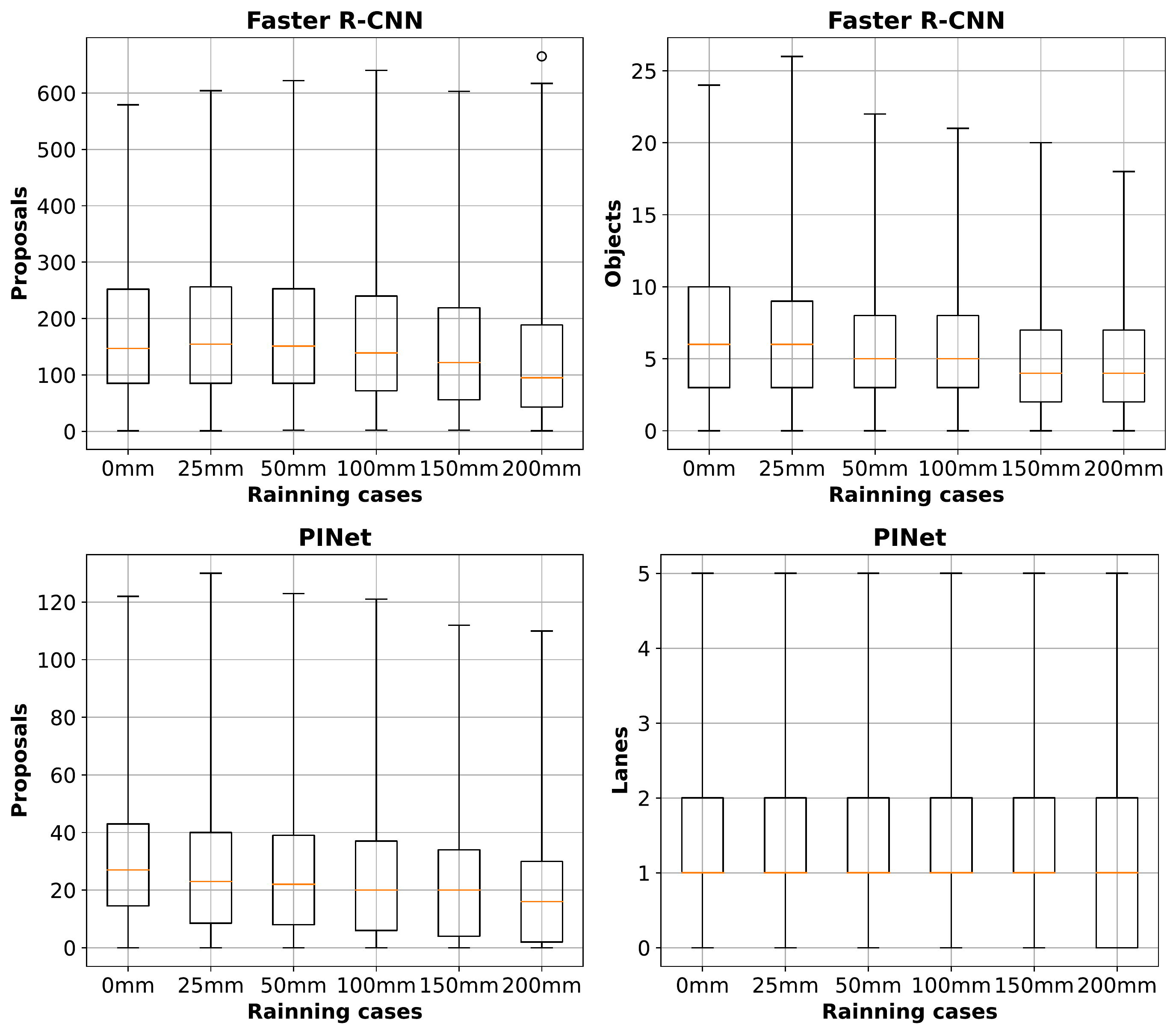}
	\caption{The box plot of proposals and objects for Faster R-CNN and PINet under different raining levels.}
	\label{fig:rain-proposal-object}
\end{figure}

\vspace{0.3em}
\noindent\textbf{Weather's impact.} When the model is deployed in a real-world environment, the weather is expected to affect the accuracy and the inference time of DNN models. To show the weather's impact on the DNN inference time variation, we render different levels of rain into the KITTI dataset and measure the end-to-end model inference time~\cite{tremblay2021rain}. Six raining cases (0/25/50/100/150/200 mm/hour) are covered, and we choose Faster R-CNN and PINet to show the impact of rains on the inference time. Table~\ref{tab:rain-e2e} shows the results of end-to-end inference time for Faster R-CNN and PINet under six rainy cases. By comparing the mean ($\mu$), the variation ($\sigma$), and the coefficient of variation ($c_{v}$) under different raining cases, we find that the end-to-end inference time's value and variation decrease as the rain level increases. The reason is that heavy rain makes thousands of pixel value changes, and the probability for a group of pixels to be lanes and objects is decreased. To prove it, we record the number of object box proposals, pixel lane proposals, detected objects, and lanes from the inference pipeline of Faster R-CNN and PINet, shown in Figure~\ref{fig:rain-proposal-object}. We can find the number of proposals for objects and lanes is decreased when the rain level increases. Besides, considering the 25th and 75th percentile of the distribution, the range and the variation of objects and lanes are also decreased, which is highly correlated with the end-to-end inference time variations. 

\noindent\textbf{Insight 1:}~\textit{The scenario affects the DNN inference time variations by the potential number of detected lanes/objects. The pixel distributions have a higher impact on lane detection than object detection. The time variations of object and lane detection decrease as the rain level increases.}

\begin{figure}[t]
 \centering
 \subfigure[\label{fig:ros1-pubsub}]{\includegraphics[width=0.5\linewidth]{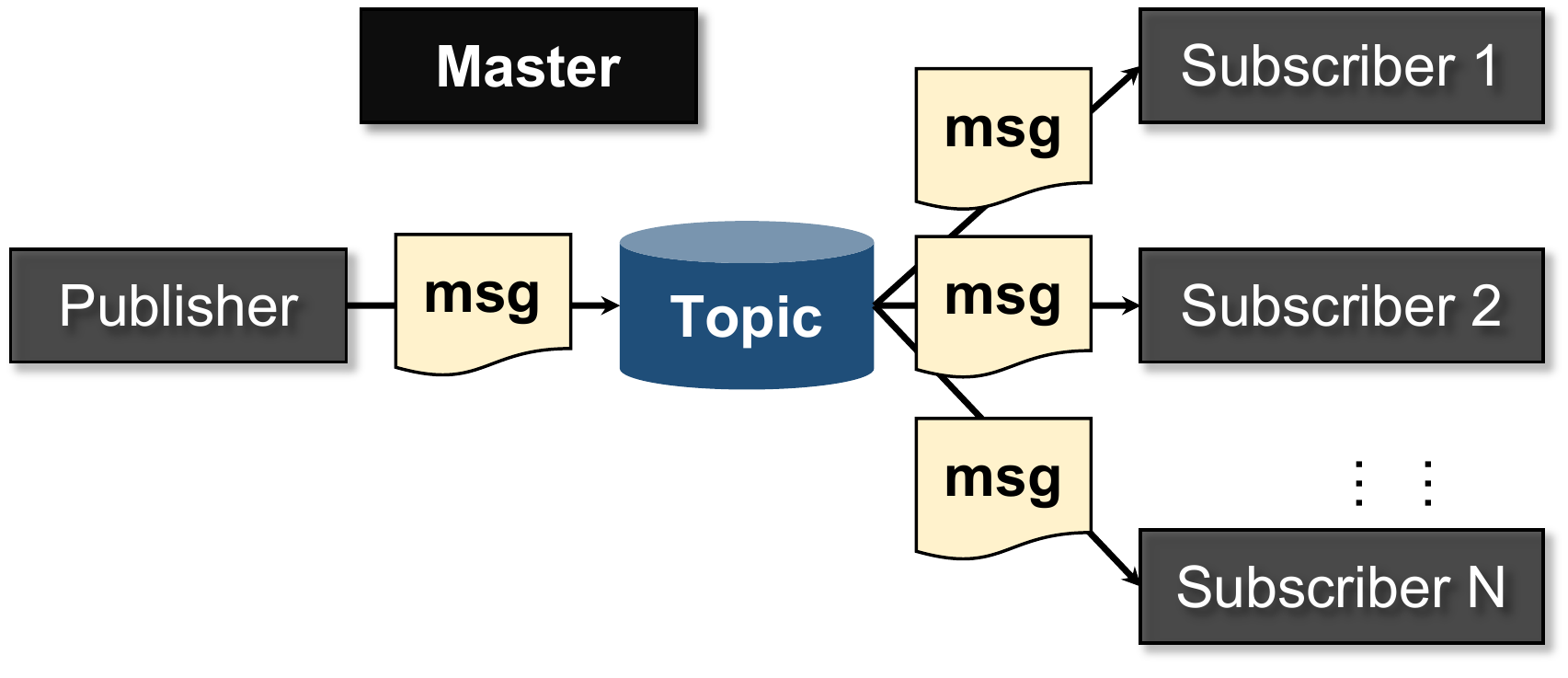}}
 \subfigure[\label{fig:ros2-pubsub}]{\includegraphics[width=0.4\linewidth]{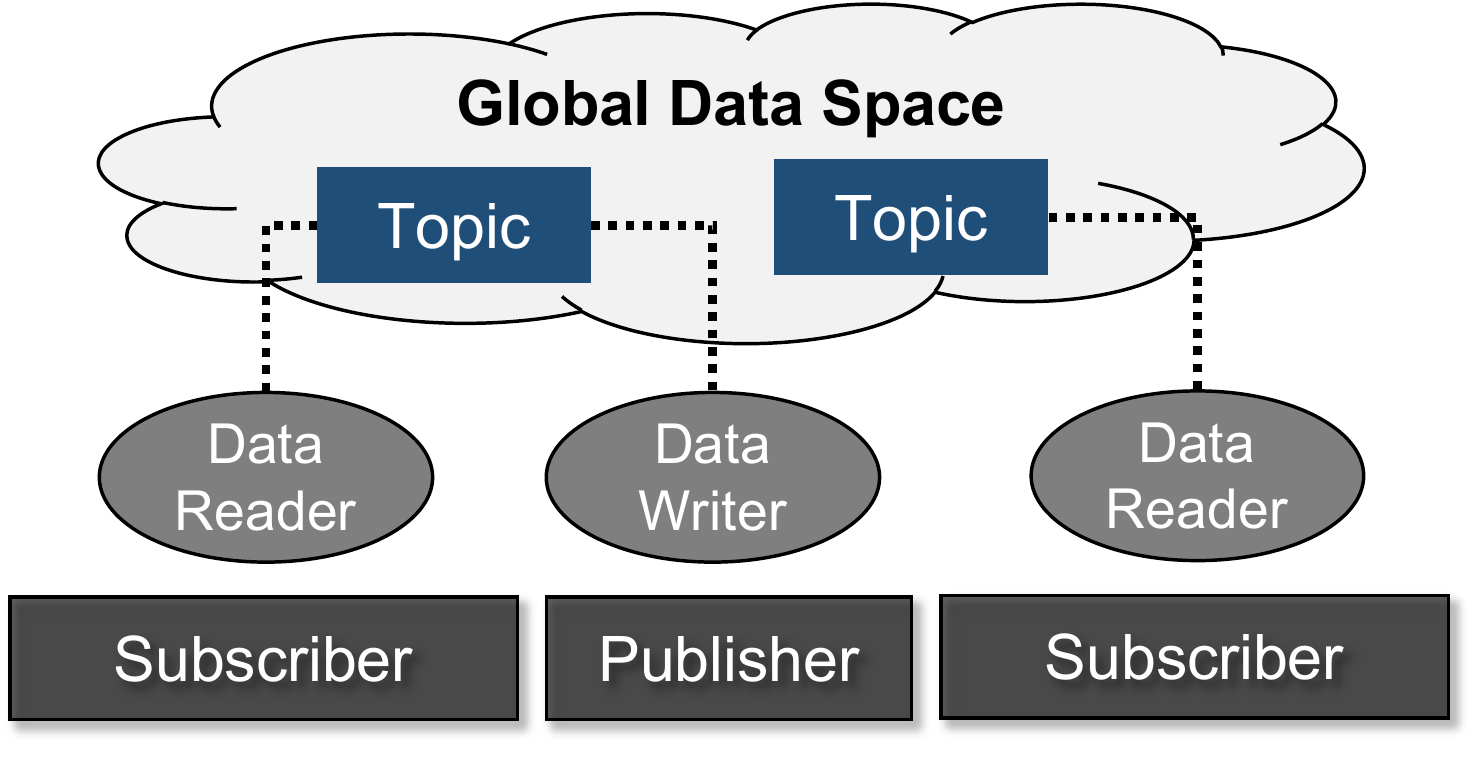}}
 \caption{The publish-subscribe model in (a) ROS1 IPC (b) ROS2 DDS. Each publisher/subscriber is corresponding to a Node in ROS1 or a participant in ROS2.}
 \label{fig:pubsub}
\end{figure}

\subsection{I/O Variability}

As an essential step for the DNN inference, access to the data could affect the whole execution pipeline. ROS is an anonymous publish-subscribe middleware system widely used for data communication. The publisher-subscriber mechanism is the main communication pattern in the autonomous driving system since multiple models need sensor data for localization, detection, segmentation, etc. The socket-based Inter-Process Communication (IPC) mechanism in ROS1 brings high compatibility and extensibility~\cite{liu2020robotic}. ROS2 is an evolved version of ROS1. ROS2 uses Data Distribution Service (DDS)~\cite{pardo2003omg} as its communication foundation, which works well in real-time distributed systems. In this part, we compare the time variations under two types of communication mechanisms: ROS IPC and ROS2 DDS.

\begin{table}
\caption{Three messages used in comparison of ROS1 IPC and ROS2 DDS.}
\label{tab:io-msg-type}
\resizebox{\linewidth}{!}{%
\begin{tabular}{|c|c|c|c|c|}
\hline
\textbf{Name} & \textbf{Message type} & \textbf{Source} & \textbf{Dimention( {[}width, height, channel{]} )} & \textbf{Size} \\ \hline
msg1 & \texttt{Image} & Image File & {[}192, 108, 3{]}   & 62KB  \\ \hline
msg2 & \texttt{Image} & Image File & {[}1920, 1080, 3{]} & 6.2MB \\ \hline
msg3 & \texttt{Image} & Camera     & {[}1920, 1080, 3{]} & 6.2MB \\ \hline
\end{tabular}
}
\end{table}

\begin{figure*}[!htp]
	\centering
	\includegraphics[width=.9\textwidth]{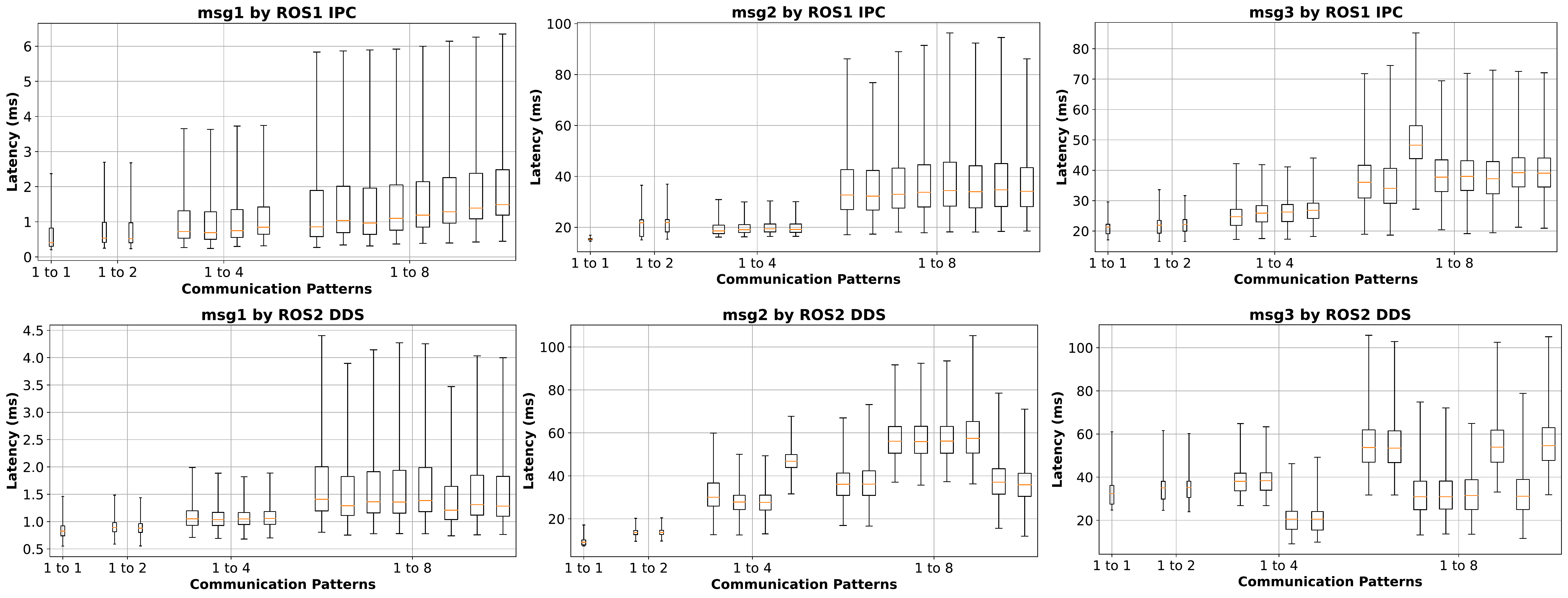}
	\caption{The box plot for message transmission latency of ROS1 IPC and ROS2 DDS when the number of subscribers increased from one to eight.}
	\label{fig:io-comparison}
\end{figure*}

Figure~\ref{fig:pubsub} shows the structure of the publish-subscribe model used in ROS1 Inter-Process Communications (IPC) and ROS2 (Data Distribution Service) DDS. ROS1 has a centralized design for data communication. It has a master node that provides naming and registration services for all the other ROS nodes, topics, services, and actions. ROS node is a process to perform a particular computation, while ROS topics are named buses for ROS nodes to exchange messages~\cite{ROS}. Figure~\ref{fig:ros1-pubsub} is a general 1 to N communication pattern (one data publisher, N data subscribers). The underlying transmission is based on \texttt{TCPROS} by default~\cite{tcpros}. When \texttt{N} nodes subscribe to a topic, the message would be copied \texttt{N-1} times and sent to the subscriber in sequence order. Unlike ROS1, ROS2 follows a distributed design for fault tolerance purposes. As is shown in Figure~\ref{fig:ros2-pubsub}, a global data space is implemented in DDS, which all independent applications can access. The underlying communication in ROS2 DDS is based on UDP and shared memory to avoid data copies~\cite{fastrtps}.

To compare the performance of ROS1 IPC and ROS2 DDS, we deployed three \texttt{Image} messages in ROS/ROS2. Table~\ref{tab:io-msg-type} shows the descriptions of these messages: one is read directly from a USB camera with a resolution of 1920x1080, another is randomly generated with the same size as the camera's frame, and the remaining one has a smaller size than the former two messages. The \texttt{Image} publisher's queue size is 1 for both ROS1 IPC and ROS2 DDS. The deployed ROS DDS is \texttt{eProsima Fast DDS}, explicitly optimized for ROS2~\cite{fastrtps,wu2021oops}. We measure the latency of message transmission from the time a message is published until the time another node subscribes to it. In addition, we set up a different number of subscribers between one to eight and recorded the communication latency for each subscriber. 

The results of the box plot for communication latency with ROS1 IPC and ROS2 DDS are shown in Figure~\ref{fig:io-comparison}. By comparing the latency distribution, we find that for both ROS1 IPC and ROS2 DDS, the range of communication latency increases when the number of subscribers increases, which indicates the I/O time variations increase when more subscribers are accessing the same \texttt{Topic}. Moreover, when the message size is small (\texttt{msg1}), ROS2 DDS shows lower communication latency and variations than ROS1 IPC. This is owing to the overhead of message copy in ROS1 IPC. However, when the message size becomes larger (\texttt{msg2}), ROS1 IPC begins to show better performance than ROS2 DDS. The reason is that ROS2 DDS invokes UDP calls for communication, while the maximum UDP datagram size in ROS2 is 64KB. Plenty of time is consumed by the message splitting and merging in ROS2 DDS, which is much higher than message copy overhead in ROS1 IPC. Besides, the communication latency variation among subscribers in ROS1 IPC is also lower than in ROS2 DDS. We can find that when transmitting \texttt{msg2} and \texttt{msg3} with one to eight patterns, four have lower latency and a smaller range, while the other four subscribers have much higher results. The main reason is that the message splitting and merging in UDP consumes so many CPU calls that the communication cannot support eight links simultaneously.

\vspace{0.3em}
\noindent\textbf{Insight 2:}~\textit{The variations of I/O latency increase significantly when the number of subscribers to the same topic increases. ROS2 DDS shows lower latency and time variations for small messages, while ROS1 IPC performs better for large messages.}

\subsection{Model Variability}

In the end-to-end timeline of DNN inference, the model plays an essential part in its time variations. Models trained under different scenarios or network structures are expected to perform differently. Within the DNN inference, most of the time is consumed by the model inference, which raises the question of whether the inference's variation also dominates the model's time variations. Since the model's variability is a complex topic, we focus on six detection models in this part.

\begin{figure}[!htp]
	\centering
	\includegraphics[width=.9\columnwidth]{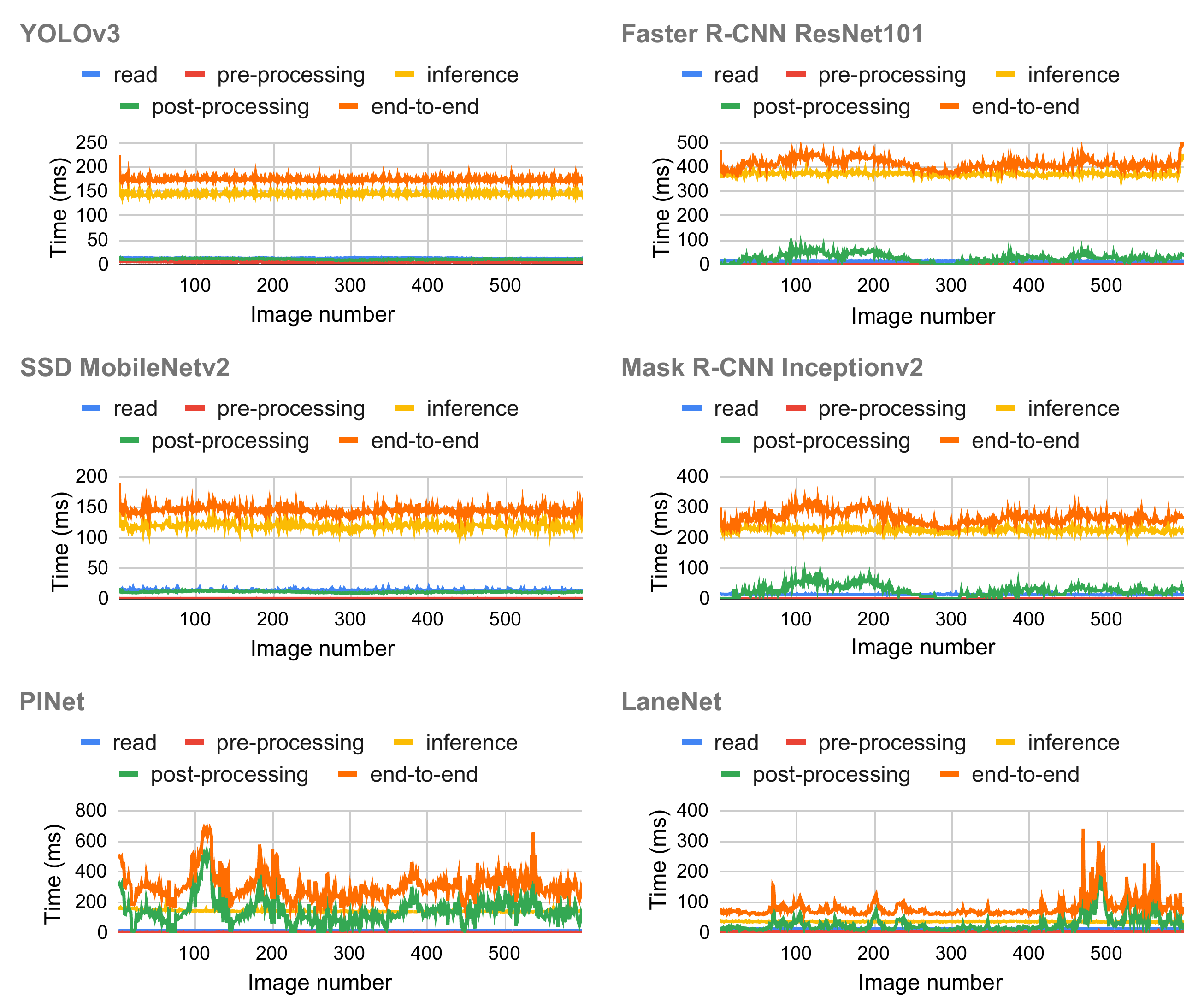}
	 \caption{The latency breakdown and time variations for four object detection models and two lane detection models with the images dataset. We can observe the relationship of the end-to-end inference time variations with the breakdown's latency. Among all these six models, the time variations of YOLOv3, SSD-MobileNetv2 are determined by model inference, while the post-processing determines the time variations of the remaining four models.}
 \label{fig:model-time-analysis}
\end{figure}

\vspace{0.3em}
\noindent\textbf{Inference and post-processing dominated.} In Figure~\ref{fig:model-time-analysis}, we apply four DNN models for object detection and two for lane detection on the image dataset to show the changing latency breakdowns with different images. Based on each part's trend in each model, we can divide the four models into two groups: inference-dominated and post-processing dominated. The inference-dominated model means the variation of the end-to-end latency is correlated with the inference time, which contains YOLOv3 and SSD-MobileNetv2. The post-processing dominated model means the variation of the end-to-end latency is greatly affected by the post-processing part. The detailed correlation analysis results for end-to-end latency with breakdowns (reading, pre-processing, inference, and post-processing) are shown in Table~\ref{tab:corr-analysis}, which gives quantitative proof of dominating factors of DNN inference time. 

Why does it happen for these DNN models? We found the original answer when looking inside the design for the DNNs. As we have learned from the design of these six models, YOLOv3 and SSD use a one-stage approach that uniformly samples on the image to get a certain number of bounding boxes and relies on the convolution layers for feature extraction to calculate the probability of object class. This one-stage design leads to a static number of objects from the inference part, enabling minor time variations of the post-processing step. In contrast, Faster R-CNN ResNet101 and Mask R-CNN Inceptionv2 are based on a two-stage approach. The first stage generates a sparse set of candidate object proposals, and the second one determines the accurate object regions and the corresponding class/lane labels using convolutional neural networks~\cite{zhang2018single}. This two-stage design causes slight time variations in the convolution neural networks. However, it makes the number of objects in post-processing dynamic, which explains why the time variations are correlated to post-processing time. 

\begin{figure}[!htp]
	\centering
	\includegraphics[width=.8\columnwidth]{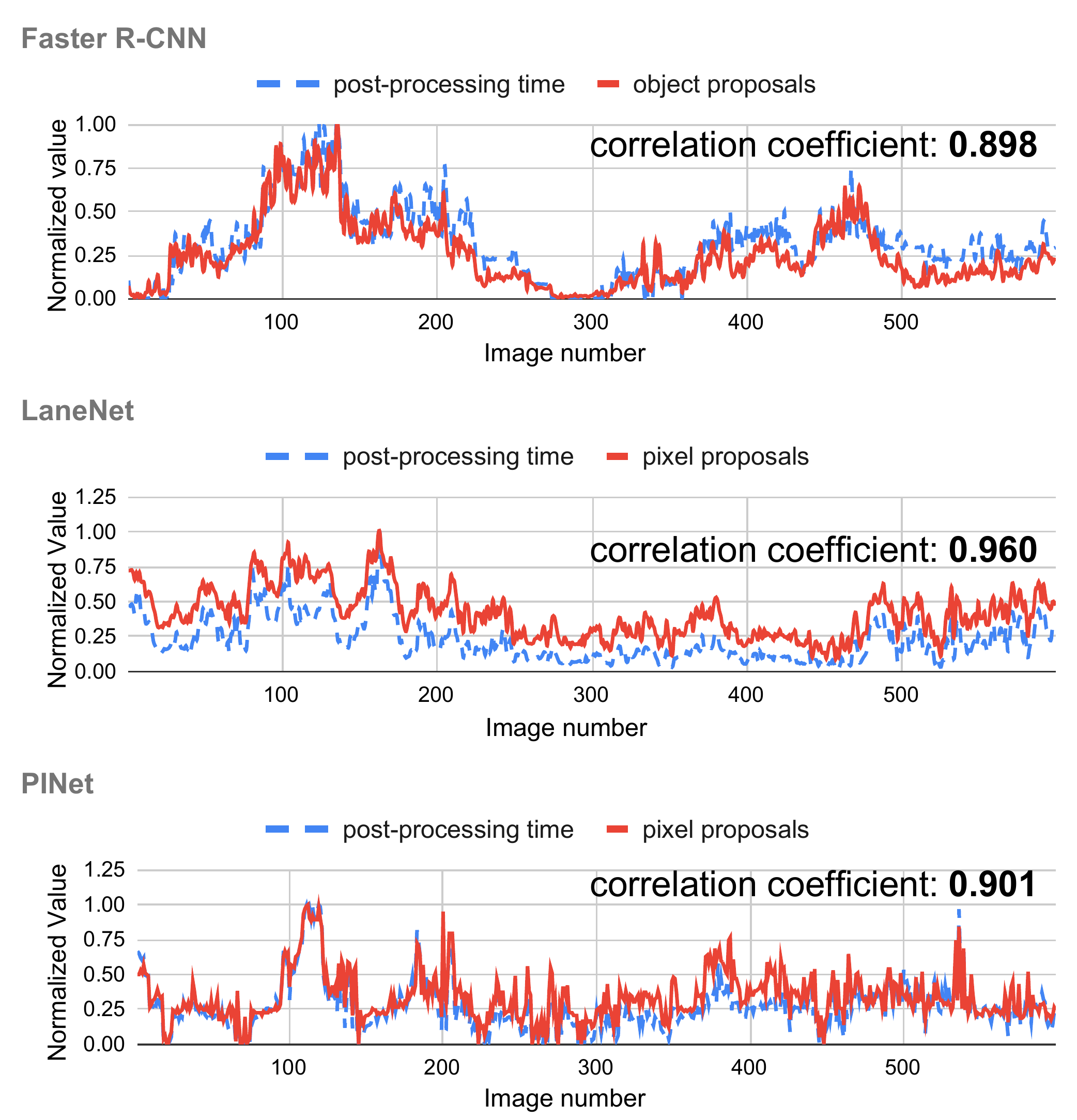}
	 \caption{The time sequence and correlation coefficients of proposals and post-processing time}
    \label{fig:proposal-post}
\end{figure}

Similarly, LaneNet and PINet also follow a two-stage design, where the first stage generates pixel proposals and the second stage clusters pixel proposals into lane groups. To prove our assumption, we collect the number of proposals and the post-processing time from DNN inference. Then we normalize them and calculate their correlation coefficients. Figure~\ref{fig:proposal-post} shows the time sequence and correlation coefficient results for Faster R-CNN, LaneNet, and PINet. The correlation coefficient between the number of proposals and post-processing time is constantly higher than 0.89.


\begin{table}
\centering
\caption{Correlation coefficients of End-to-End latency with the breakdowns. The more closer to 1, the more correlated.}
\resizebox{\linewidth}{!}{%
\begin{tabular}{ccccc} 
\toprule
\textbf{Correlation Coefficients} & \textbf{read} & \textbf{pre-processing} & \textbf{inference} & \textbf{post-processing} \\ 
\hline
\textit{YOLOv3} & 0.220 & 0.429 & \textbf{0.906} & 0.378 \\
\textit{Faster R-CNN ResNet101} & -0.108 & 0.060 & 0.681 & \textbf{0.896} \\
\textit{Mask R-CNN Inceptionv2} & -0.131 & 0.035 & 0.619 & \textbf{0.946} \\
\textit{SSD MobileNetv2} & 0.169 & 0.056 & \textbf{0.963} & 0.525 \\
\bottomrule
\end{tabular}
\label{tab:corr-analysis}
}
\end{table}

\vspace{0.3em}
\noindent\textbf{Insight 3:}~\textit{The design of the model's structure significantly impacts the time variations of the DNN inference. For object detection, one-stage models show less variation than two-stage based ones. The time variations for two-stage based object detection and lane detection models are mainly caused by the number of objects/lanes proposals from the first stage.}

\subsection{Runtime Variability}

With fixed input, model, and hardware, runtime becomes an essential factor contributing to the time variation of DNN inference. However, since a simple application can invoke thousands of system calls with millions of instructions to computing architecture, it is tough to accurately predict the execution time for a given application. Since GPU tasks are non-preemptive, CPU scheduling becomes the main uncertainty in runtime~\cite{basaran2012supporting}. In this paper, the DNN inference runtime profiling focuses on finding the connections between the CPU scheduling policy and the time variation.

\vspace{0.3em}
\noindent\textbf{Scheduling policy setup.}
Due to the hard real-time requirements from safety-critical applications, real-time operating systems or systems with real-time kernel patches are widely used in computing systems for autonomous vehicles. In this paper, we use the NVIDIA Jetson AGX board and configure it with an RT-kernel patch. In \textit{L4T R32.4.2}, the RT kernel allows preemption for most system calls~\cite{reghenzani2019real}, giving more space to guarantee the deadlines for safety-critical applications.

We choose four scheduling policies: \textit{SCHED\_OTHER}, \textit{SCHED\_FIFO}, \textit{SCHED\_RR}, and \textit{SCHED\_DEADLINE}. \textit{SCHED\_OTHER} is AGX's default scheduling policy for user applications for maximum processor utilization. \textit{SCHED\_FIFO} schedules in a first-come-first-serve method, while \textit{SCHED\_RR} schedules in a round-robin way. \textit{SCHED\_DEADLINE} is a CPU scheduler based on the Earliest Deadline First (EDF)~\cite{spuri1994efficient}. We choose two DNN models, PINet and YOLOv3, for the runtime profiling. The priorities for PINet and YOLOv3 are shown in Table~\ref{tab:priority-setup}. It shows that the priorities of \textit{SCHED\_FIFO} and \textit{SCHED\_RR} are all 99, while others are 0. For the setting up of deadlines, we choose two deadlines for each of them. Deadline-1 is set up based on the worst observed execution time (225ms for YOLOv3 and 300ms for PINet). Deadline-2 is based on the average end-to-end DNN inference time (200ms for YOLOv3 and 150ms for PINet). Under each scheduling policy, the mean value, 50, 80, and 99 percentiles, is used to show end-to-end latency distribution. Besides, the coefficient of variation is also calculated to show the quantified time variations.

Since the competition of applications significantly impacts the execution time, we conduct experiments for each model in two steps. First, we run the model inference without competition, called the single test. Then we conduct the experiments with the resource competition from another DNN model, called the compete test. By default, the competition model will use \textit{SCHED\_OTHER} as the scheduling policy.

\begin{table}
\centering
\caption{The scheduling setup for PINet and YOLOv3.}
\label{tab:priority-setup}
\resizebox{\linewidth}{!}{%
\begin{tabular}{cccc} 
\toprule
\textbf{ Scheduling Policies\textit{ }} & \begin{tabular}[c]{@{}c@{}}\textbf{Priority}\\\textbf{min/max }\end{tabular} & \textbf{PINet } & \textbf{YOLOv3 } \\ 
\hline
\textcolor[rgb]{0.137,0.161,0.18}{\textit{SCHED\_OTHER}}  & 0/0 & 0 & 0 \\
\textcolor[rgb]{0.137,0.161,0.18}{\textit{SCHED\_FIFO}}  & 1/99 & 99 & 99 \\
\textcolor[rgb]{0.137,0.161,0.18}{\textit{SCHED\_RR}}  & 1/99 & 99 & 99 \\ 
\hline
\textcolor[rgb]{0.137,0.161,0.18}{\textit{SCHED\_DEADLINE}}  & 0/0 & \begin{tabular}[c]{@{}c@{}}0\\deadline-1: 300ms\\deadline-2: 150ms \end{tabular} & \begin{tabular}[c]{@{}c@{}}0\\deadline-1: 225ms\\deadline-2: 200ms \end{tabular} \\
\bottomrule
\end{tabular}
}
\end{table}

\begin{figure}[!htp]
	\centering
	\includegraphics[width=\columnwidth]{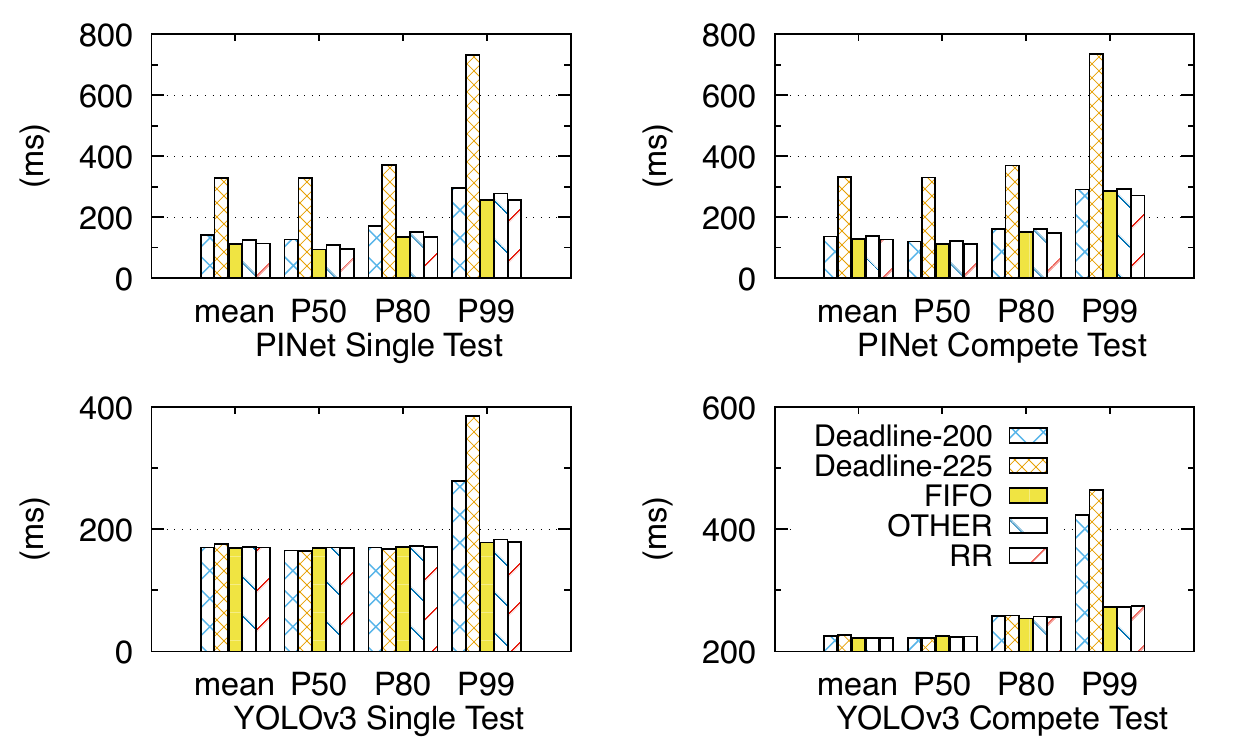}
	 \caption{Latency's mean, 50, 80, and 99 percentiles of PINet and YOLOv3.}
    \label{fig:runtime-analysis-rt}
\end{figure}

\vspace{0.3em}
\noindent\textbf{Time variations under RT kernel.}
The RT kernel results in latency’s mean, 50, 80, and 99 percentiles are shown in Figure~\ref{fig:runtime-analysis-rt}. 
Deadline-based scheduling shows the worst time variations among all the RT scheduling policies (FIFO, RR, and Deadline). One explanation is that the scheduler does not terminate tasks even when it has already passed the deadline. The variation of the deadline-based approach can be decreased with the termination, but it will reduce the FPS of the detection. Besides, deadline-based scheduling with the average time is much better than the deadline with the worst observed execution time, which raises another question in selecting an appropriate deadline. The quantified results for the coefficient of variation are shown in Table~\ref{tab:cv-rt}. The results prove our analysis above that the deadline-based scheduling shows the worst performance in variation. Besides, the DNN inference time variations are much more severe in the compete test than in the single test, mainly caused by contention for non-preemptive system resources. Analysis of the time variations for multi-tenant DNN inference exceeds the topic of this paper and will be our future work.

\vspace{0.3em}
\noindent\textbf{Insight 4:}~\textit{Deadline-based scheduling shows more time variations than other real-time scheduling policies. Setting average time as the deadline has much fewer time variations than the worst observed execution time in DNN inference.}



\begin{table}
\centering
\caption{Coefficient of variation under RT kernel.}
\resizebox{\linewidth}{!}{%
\begin{tabular}{ccccc} 
\toprule
 \textbf{$c_{v}$ - RT}  & \textbf{PINet-single}  & \textbf{PINet-compete}  & \textbf{YOLOv3-single}  & \textbf{YOLOv3-compete}  \\ 
\hline
\textit{Deadline-1}  & 0.30 & 0.33 & 0.13 & 0.21 \\
\textit{Deadline-2}  & \textbf{0.44 } & \textbf{0.44 } & \textbf{0.27} & \textbf{0.23 } \\
\textit{FIFO}  & 0.33 & 0.31 & 0.02 & 0.14 \\
\textit{OTHER}  & 0.32 & 0.30 & 0.02 & 0.15 \\
\textit{RR}  & 0.33 & 0.31 & 0.02 & 0.15 \\
\bottomrule
\end{tabular}
\label{tab:cv-rt}
}
\end{table}

\subsection{Hardware Variability}

Since most autonomous driving applications are DNN-based and require massive computations, the variability of the hardware in CPU, GPU, and memory configurations dramatically affects the DNN inference performance. This part discusses the hardware's variability in two aspects: end-to-end latency for different devices and time variations under CPU and GPU-based architectures.



\begin{figure}[!htp]
	\centering
	\includegraphics[width=.9\columnwidth]{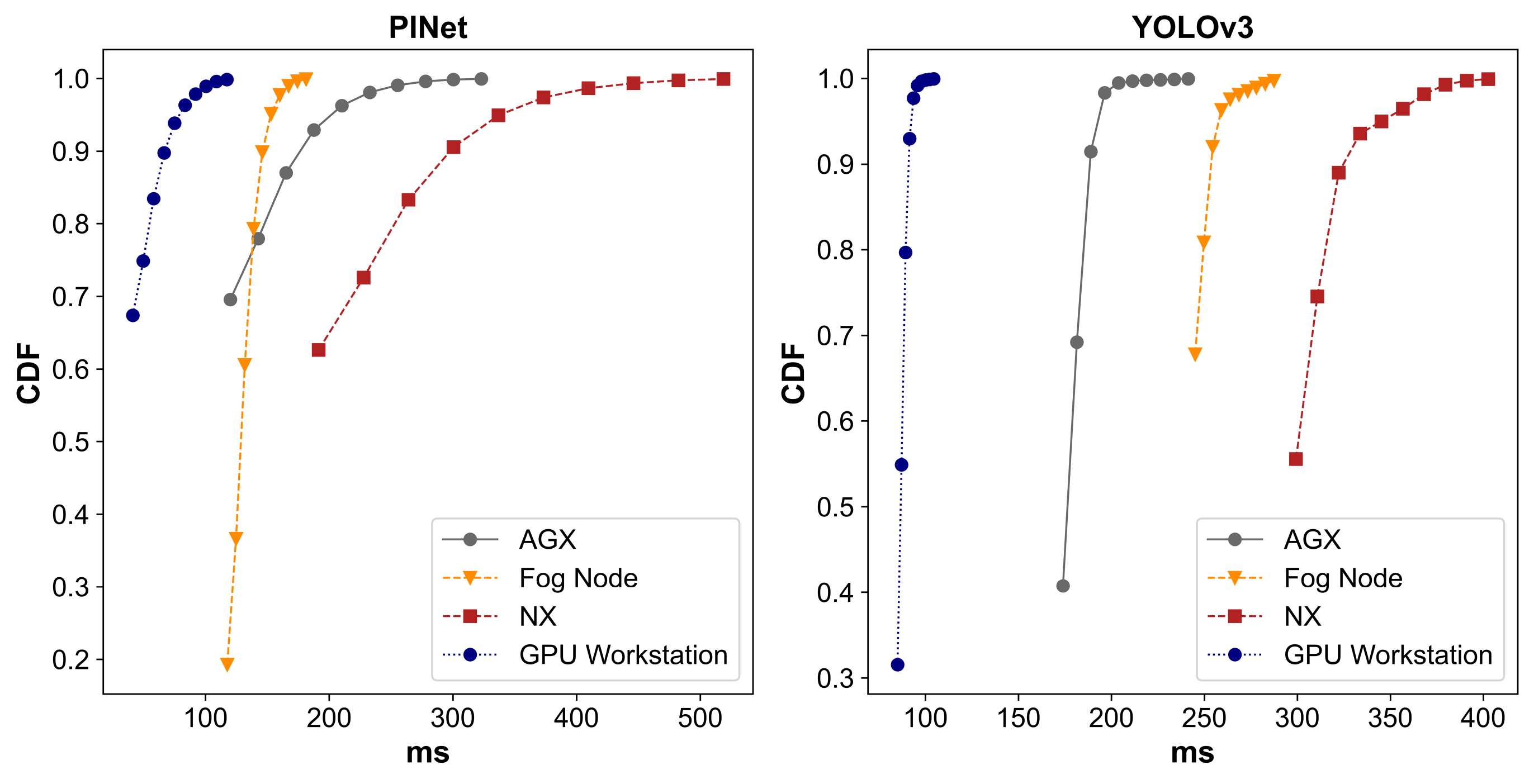}
	\caption{\major{The CDF of end-to-end latency for PINet and YOLOv3 with different hardware.}}
	\label{fig:hardware-cdf}
\end{figure}

\vspace{0.3em}
\noindent\textbf{End-to-end latency for different devices.} We use all four devices listed in Table~\ref{tab:hardware-description} for running PINet and YOLOv3 model inference experiments to collect the end-to-end latency with breakdowns. The results are shown in Figure~\ref{fig:hardware-cdf}. \major{The GPU workstation performs the best in average end-to-end inference latency among all the devices. However, all devices show long tail latency with non-negligible time variations. For two-stage based PINet, the Fog Node tends to have a smaller time variation range than Jetson AGX and Xavier NX. The reason is that the post-processing mainly causes the time variations of PINet on the CPU side, while the Fog node has more powerful CPUs. For YOLOv3, which is one-stage based, the GPU desktop shows the lowest inference time variations because the inference part mainly dominates the time variations of the one-stage based model on the GPU side. The GPU workstation has much higher AI performance in TOPs than the other two Jetson boards.} 

\vspace{0.3em}
\noindent\textbf{Time variations under CPU \& GPU architecture.} Since the average latency of PINet and YOLOv3 on Jetson AGX, Fog Node, and GPU workstation are close. At the same time, the inference time variations have a big difference. We use these three devices to profile the impact of CPU and GPU architectures on DNN time variations. To begin with, we use \textit{nvprof} to profile the GPU activities on AGX Xavier. We found that for PINet and YOLOv3, only the inference part is executed on GPU while read, pre-processing, and post-processing are usually executed on the CPU. To find the lower level roots causing the time variations in post-processing, we use \textit{Linux perf} to monitor the system events when PINet and YOLOv3 models are executed with/without post-processing. The collected system events include the instructions, cache references, cache misses, CPU cycles, context switches, CPU migrations, etc.
Table~\ref{tab:hardware-perf} shows the ratio of collected events on DNN inference between with and without post-processing. The higher the value is, the more events happen during the post-processing. Based on the comparison results, we can find that more events happen in AGX Xavier than Fog node, including cache misses, instructions, cycles, context switches, CPU clocks, and CPU migrations. Events like CPU migrations, context switches, cycles, and CPU clocks directly contributed to the time variations in post-processing. The Fog node has a server-level CPU, while the AGX Xavier's CPU is mainly used on embedding and mobile systems. We conclude that embedding systems on autonomous vehicles with more powerful CPUs could help decrease the time variations in DNN inference. 

\begin{table}[]
\caption{The events comparison of running PINet and YOLOv3 on Jetson AGX and Fog Node with/without post-processing.}
\label{tab:hardware-perf}
\resizebox{\columnwidth}{!}{%
\begin{tabular}{|l|ccc||ccc|}
\hline
\textit{w. / w. o. post-process} & \multicolumn{3}{c||}{\textbf{PINet}}                                                    & \multicolumn{3}{c|}{\textbf{YOLOv3}}                                                  \\ \hline
\textit{metrics}                 & \multicolumn{1}{c|}{\textbf{AGX}}  & \multicolumn{1}{c|}{\textbf{Fog}} & \textbf{GPU}  & \multicolumn{1}{c|}{\textbf{AGX}}  & \multicolumn{1}{c|}{\textbf{Fog}} & \textbf{GPU} \\ \hline
\hline
\textit{branch-misses}           & \multicolumn{1}{c|}{\textbf{1.18}} & \multicolumn{1}{c|}{1.02}         & 1.07          & \multicolumn{1}{c|}{1}             & \multicolumn{1}{c|}{1.02}         & 1.02         \\ \hline
\textit{cache-misses}            & \multicolumn{1}{c|}{1.04}          & \multicolumn{1}{c|}{1.03}         & \textbf{1.25} & \multicolumn{1}{c|}{1.03}          & \multicolumn{1}{c|}{1.03}         & 1.03         \\ \hline
\textit{cache-references}        & \multicolumn{1}{c|}{\textbf{1.09}} & \multicolumn{1}{c|}{1.03}         & 1.07          & \multicolumn{1}{c|}{\textbf{1.03}}          & \multicolumn{1}{c|}{1}            & 1.01         \\ \hline
\textit{instructions}            & \multicolumn{1}{c|}{1.06}          & \multicolumn{1}{c|}{1.04}         & \textbf{1.11} & \multicolumn{1}{c|}{\textbf{1.04}}          & \multicolumn{1}{c|}{1}            & 1            \\ \hline
\textit{cycles}                  & \multicolumn{1}{c|}{\textbf{1.12}} & \multicolumn{1}{c|}{1.02}         & 1.04          & \multicolumn{1}{c|}{\textbf{1.04}}          & \multicolumn{1}{c|}{1}            & 1.02         \\ \hline
\textit{context-switches}        & \multicolumn{1}{c|}{1.06}          & \multicolumn{1}{c|}{0.31}         & \textbf{1.15} & \multicolumn{1}{c|}{\textbf{1.36}} & \multicolumn{1}{c|}{0.99}         & 1.15         \\ \hline
\textit{cpu-migrations}          & \multicolumn{1}{c|}{0.95}          & \multicolumn{1}{c|}{0.66}         & \textbf{4.54} & \multicolumn{1}{c|}{\textbf{1.10}}          & \multicolumn{1}{c|}{1.01}         & 1.03         \\ \hline
\textit{page-faults}             & \multicolumn{1}{c|}{1}             & \multicolumn{1}{c|}{0.65}         & 1             & \multicolumn{1}{c|}{1}             & \multicolumn{1}{c|}{1.04}         & 0.98         \\ \hline
\end{tabular}%
}
\end{table}


\vspace{0.3em}
\noindent\textbf{Insight 5:}~\textit{The time variations in PINet are mainly caused by the post-processing on the CPU. Although GPU performs better in conducting matrix operations, having a powerful CPU in the autonomous vehicle embedding system would help to decrease the time variations in DNN inference.}

\section{End-to-End System for Autonomous Driving}
\label{implementation}

Autonomous driving vehicles are composed of a variety of applications for sensing, perception, and control. The performance of the system is expected to rely on the coordination of several modules. To evaluate autonomous driving system's variation, we propose to build an end-to-end prototype based on ROS in the NVIDIA Jetson AGX board with RT kernel.  

\subsection{Overview of the End-to-End System}

\begin{figure}[!htp]
	\centering
	\includegraphics[width=\columnwidth]{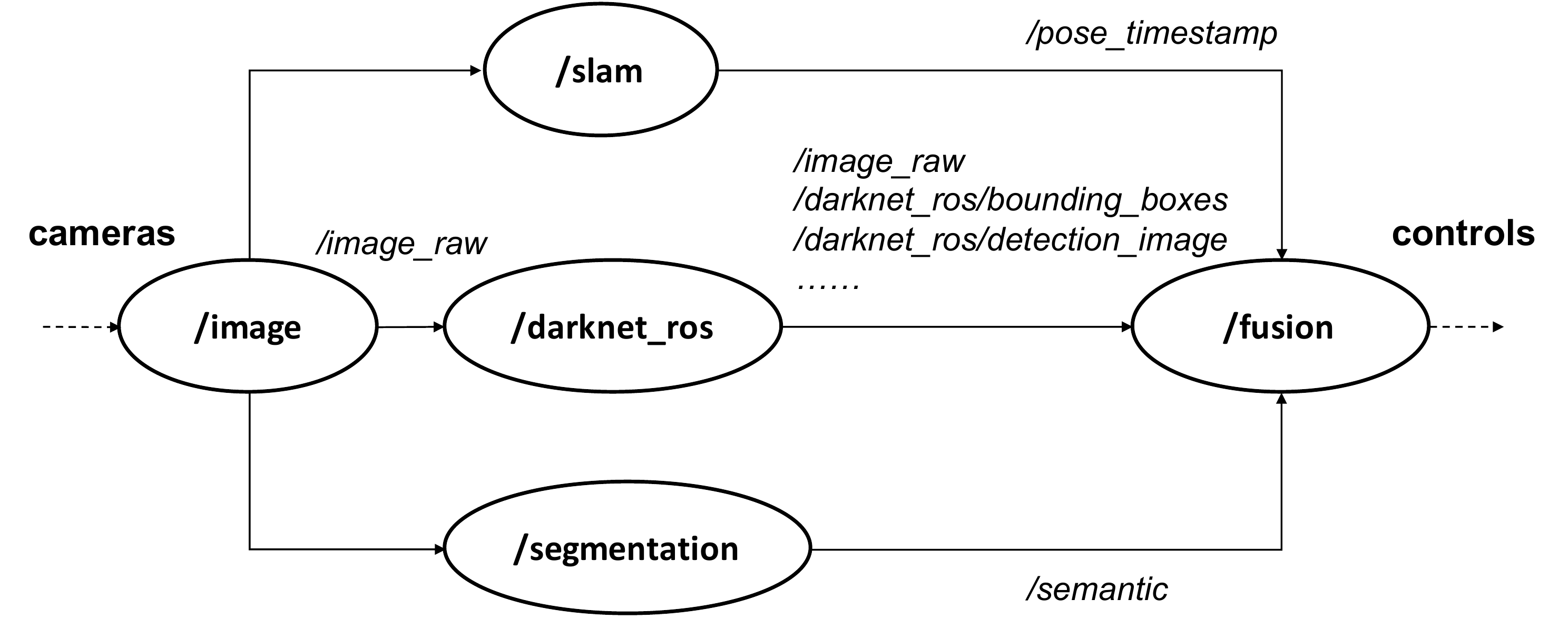}
	\caption{The ROS framework of the end-to-end system.}
	\label{fig:ros-framework}
\end{figure}

Based on the overview autonomous driving system in Figure~\ref{fig:AD-system-overview}, we develop a ROS framework for the end-to-end system, as shown in Figure~\ref{fig:ros-framework}. Since this paper focuses on the variations of DNN inference and the planning modules are all rule-based algorithms with stable execution time, the ROS framework does not include that part. The whole pipeline starts with the \textit{/image} node, capturing images from the cameras or image files, and publishing it. Three perception nodes subscribe \textit{/image\_raw} messages and execute the DNN inference on the images. Three perception nodes are responsible for Simultaneous Localization and Mapping (SLAM), object detection, and semantic segmentation. After DNN inference, three nodes publish their results, covering the position information, objects, and semantics. Another node called \textit{/fusion} subscribes to these three nodes. It synchronizes them to get the sensor fusion results, which gives the control module the location of the vehicle and obstacles and open space for driving.

In our design, the algorithm used for SLAM is ORB-SLAM2, which is a pure camera-based approach to capture key points in pixels and localize the vehicles and generate maps simultaneously~\cite{mur2016orb}. YOLOv3 is used for object detection, while Deeplabv3 is used for semantic segmentation~\cite{redmon2018yolov3,chen2017rethinking}.

\subsection{ROS Nodes and Topics}

A ROS node (i.e., the nodes in the figure) is a process to perform a particular computation while ROS topics (i.e., the arrows in the figure) are named buses for ROS nodes to exchange messages~\cite{ROS}. In the end-to-end system, we implement five ROS nodes to access the sensor data and process the data: \textit{/image}, \textit{/darknet\_ros}, \textit{/slam}, \textit{/segmentation}, and \textit{/fusion}. Publish-subscriber-based message sharing is used to transmit messages between these nodes.

The ROS topics are defined to exchange messages between ROS nodes. Ten ROS topics are implemented to exchange messages, including images, positions, objects, and other customized messages. The summarized descriptions of some ROS topics are reported in Table~\ref{tab:ros-topic}. There are two messages based on \texttt{Image} type: header, height, width, encoding, data, etc. The ROS topic's header contains the sequence ID, timestamp, and frame ID to represent a specific message. Since timestamp and sequence ID are needed for the synchronization, we implement \textit{/pose\_timestamp} based on \textit{/pose}, which contains the position and orientation data. For object detection, bounding boxes are used to present the detected objects, which is determined by min and max values of the x and y-axis and the probability and the object class. \textit{/bounding\_boxes} includes all the bounding boxes for one image and contains a header inside. For semantic segmentation, the results are shown as different colors inside the image to represent different segments. An \texttt{Image}-based topic called \textit{/semantic} is used to represent the results with the message header.

\begin{table}
\centering
\caption{ROS topics used in the end-to-end system}
\resizebox{\linewidth}{!}{%
\begin{tabular}{|c|c|c|c|} 
\hline
\textbf{ROS Topics} & \textbf{Library} & \textbf{Type} & \textbf{Fields} \\ 
\hline
\textit{/image\_raw} & sensor\_msgs & Image & \begin{tabular}[c]{@{}c@{}}header, height, width, encoding, data, etc\end{tabular} \\ 
\hline
\textit{/pose\_timestamp} & geometry\_msgs & PoseStamped & header, pose \\ 
\hline
\textit{/pose} & geometry\_msgs & Pose & \begin{tabular}[c]{@{}c@{}}position (x, y, z float64), \\orientation (x, y, z, w, float64)\end{tabular} \\ 
\hline
\textit{/bounding\_boxes} & darknet\_ros\_msgs & BoundingBoxes & \begin{tabular}[c]{@{}c@{}}header, image\_header, bounding\_boxes\end{tabular} \\ 
\hline
\textit{/bounding\_box} & darknet\_ros\_msgs & BoundingBox & \begin{tabular}[c]{@{}c@{}}probability, xmin, ymin, xmax, ymax, id, class\end{tabular} \\ 
\hline
\textit{/semantics} & sensor\_msgs & Image & \begin{tabular}[c]{@{}c@{}}header, height, width, encoding, data, etc\end{tabular} \\
\hline
\end{tabular}
\label{tab:ros-topic}
}
\end{table}

\subsection{Message Synchronization}

For ROS nodes that need to subscribe to multiple ROS topics and process them together, message synchronization becomes one of the implementation issue. Typically, message synchronization is based on the timestamp and sequence ID, so we convert the \textit{/pose} message to \textit{/pose\_timestamp} to add on the header.

The \textit{/fusion} node's objective is to combine all the perception results of the same image frame. The first thing is to make a unique ID for each image frame. In the beginning, the \textit{/image} node attaches timestamp information and frame ID to each message it publishes out. For three perception nodes, after DNN inference on the coming image frame, the timestamp and sequence ID of the coming images will be used as the header's timestamp and sequence ID of the new message like \textit{/pose\_timestamp}, \textit{/semantic}, etc. With unique IDs on each image frame and detection results, the remaining question is how to make them synchronized. In our design, we use \textit{message\_filter\footnote{http://wiki.ros.org/message\_filters}} with Approximate Time Synchronizer to manage the fusion process. The approximate synchronizer sets queue size as 100 and 100ms as the slop, which means the message with a time difference less than 100ms is considered synchronized.

\section{System-level Profiling}
\label{system-analysis}



Compared with the typical model inference, the end-to-end system is expected to have more uncertainties, contributing to higher variations. In this section, with the ROS-based end-to-end system, we investigate the time variation issues for the real autonomous driving system in two aspects: typical module time analysis and end-to-end system time analysis.

\subsection{Latency Analysis for each Module}

\begin{figure}[!htp]
	\centering
	\includegraphics[width=\columnwidth]{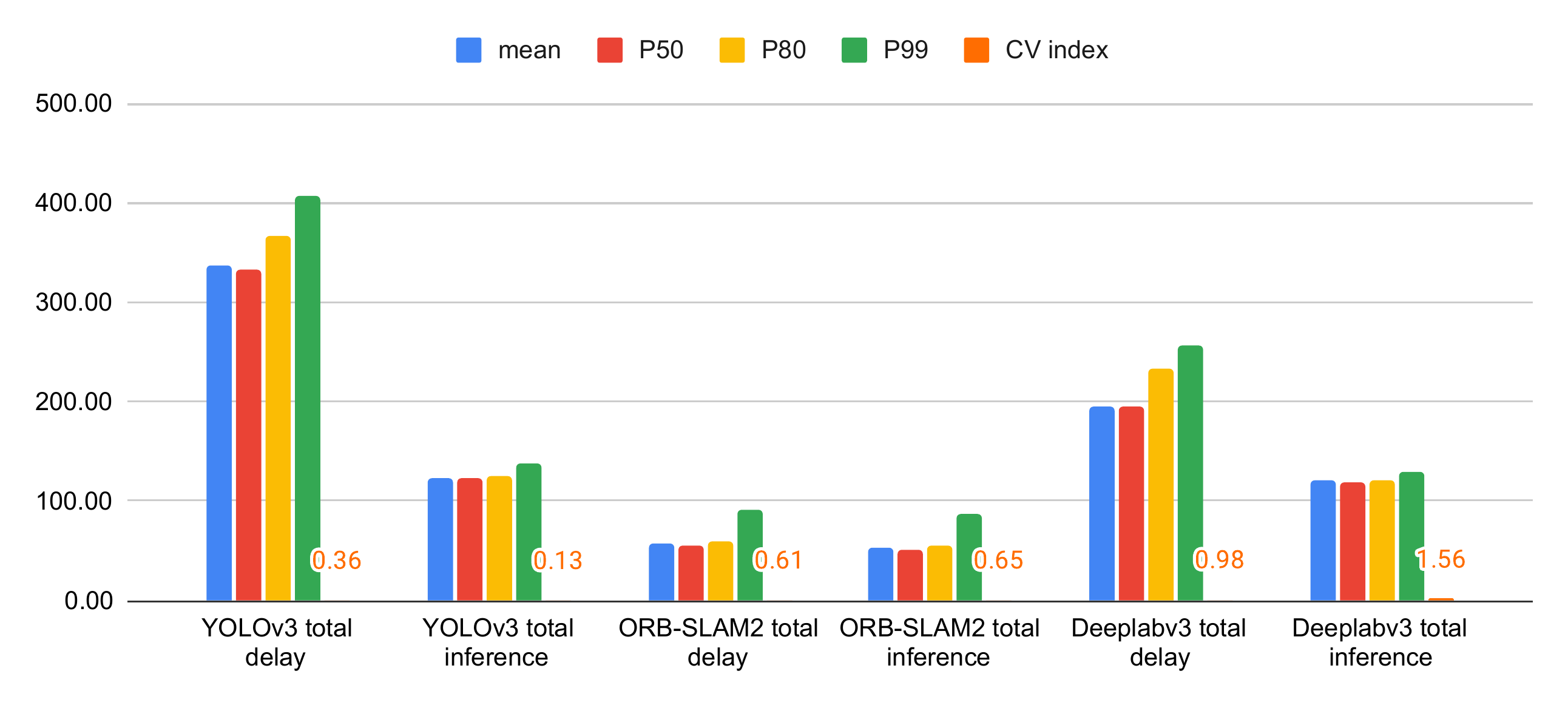}
	\caption{The latency analysis for each ROS perception module when running separately.}
	\label{fig:ros-single-module}
\end{figure}

There are three typical modules in the ROS-based prototype: ORB-SLAM2 based localization and mapping, YOLOv3 based object detection, and Deeplabv3 for semantic segmentation. These modules subscribe to the image topic from the \textit{/image} node and execute the DNN inference, which means the I/O changes into ROS communications. The total delay is defined as the difference between the model inference finish time and the image's sending timestamp from \textit{/image} node. In contrast, the total inference time is defined as the total time spent within the perception module. The results for running each module separately is shown in Figure~\ref{fig:ros-single-module}. We can observe that for YOLOv3 and Deeplabv3, the difference between total delay and total inference is huge, while ORB-SLAM2 is small. The difference is because ROS communication has overhead in transmitting images using pub/sub mechanisms. It is supposed to convert RGB images into ROS images before publishing out and needs to convert it back to RGB images for DNN inference after receiving it. The results of $C_{v}$ present that the total delay variation is higher than the total inference for YOLOv3 and Deeplabv3.

\subsection{System Latency Analysis}

\begin{figure}[!htp]
	\centering
	\includegraphics[width=\columnwidth]{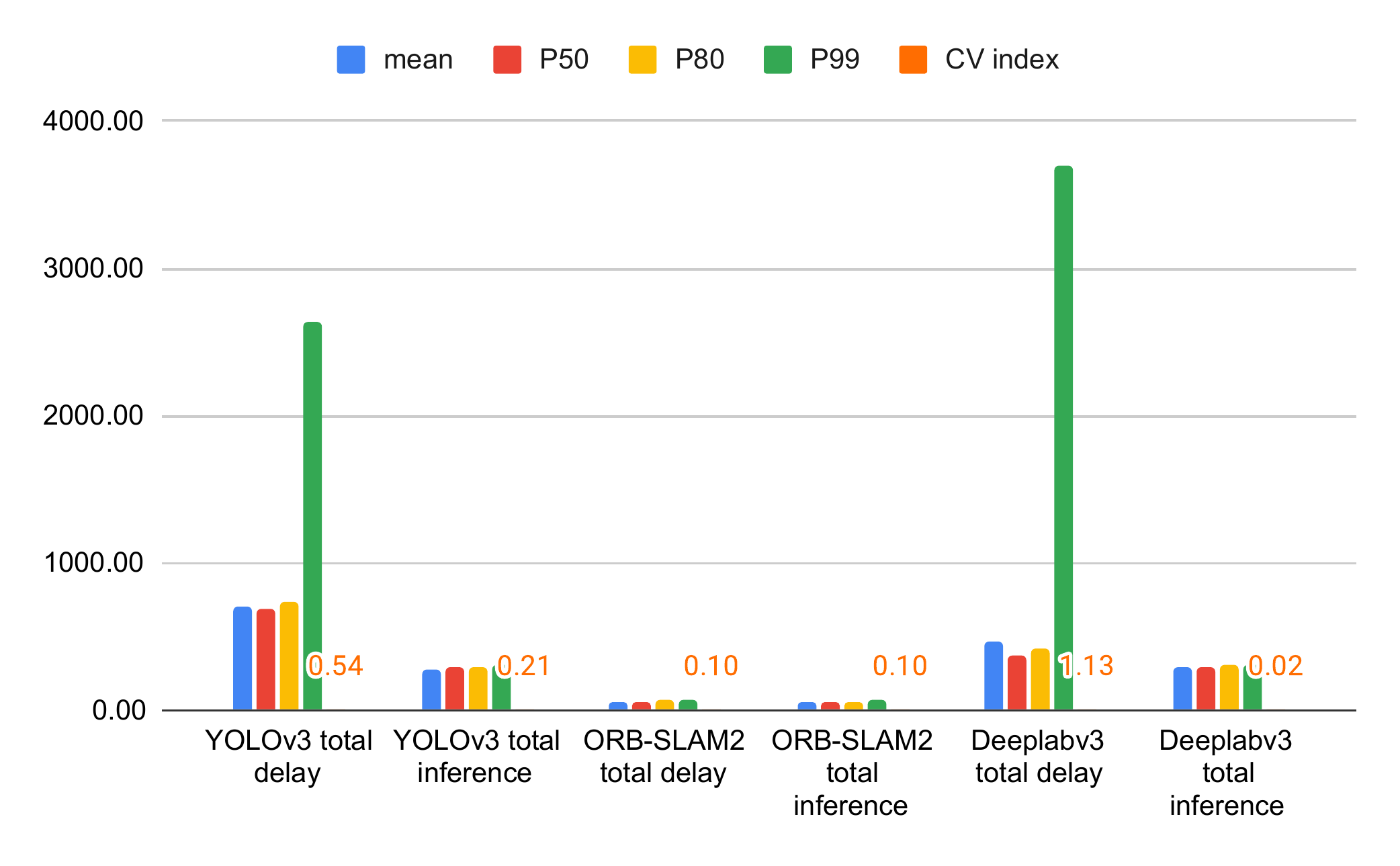}
	\caption{The latency analysis for each ROS module when the end-to-end system is running.}
	\label{fig:ros-system-module}
\end{figure}

With multiple perception modules subscribing to the \textit{/image\_raw} topics simultaneously and the ROS node \textit{/fusion} combines their results, the system latency is expected to show more variations than the separate module case. The results for the system latency of each module is shown in Figure~\ref{fig:ros-system-module}. We can find that the time variations of YOLOv3 and Deeplabv3's total delay are much higher than the typical module case - with the highest delay almost attains 4000ms, which is dangerous for the safety-critical autonomous driving systems. Besides, the 99 percentile for YOLOv3 and Deeplabv3 is more than the combination of that in a typical module case, which implies huge tail latency caused by the accumulation of variations with the competing of concurrent tasks. $C_{v}$ shows models' variations in which ORB-SLAM2 shows low variation while two DNN-based module show huge variations.

\begin{figure}[!htp]
	\centering
	\includegraphics[width=\columnwidth]{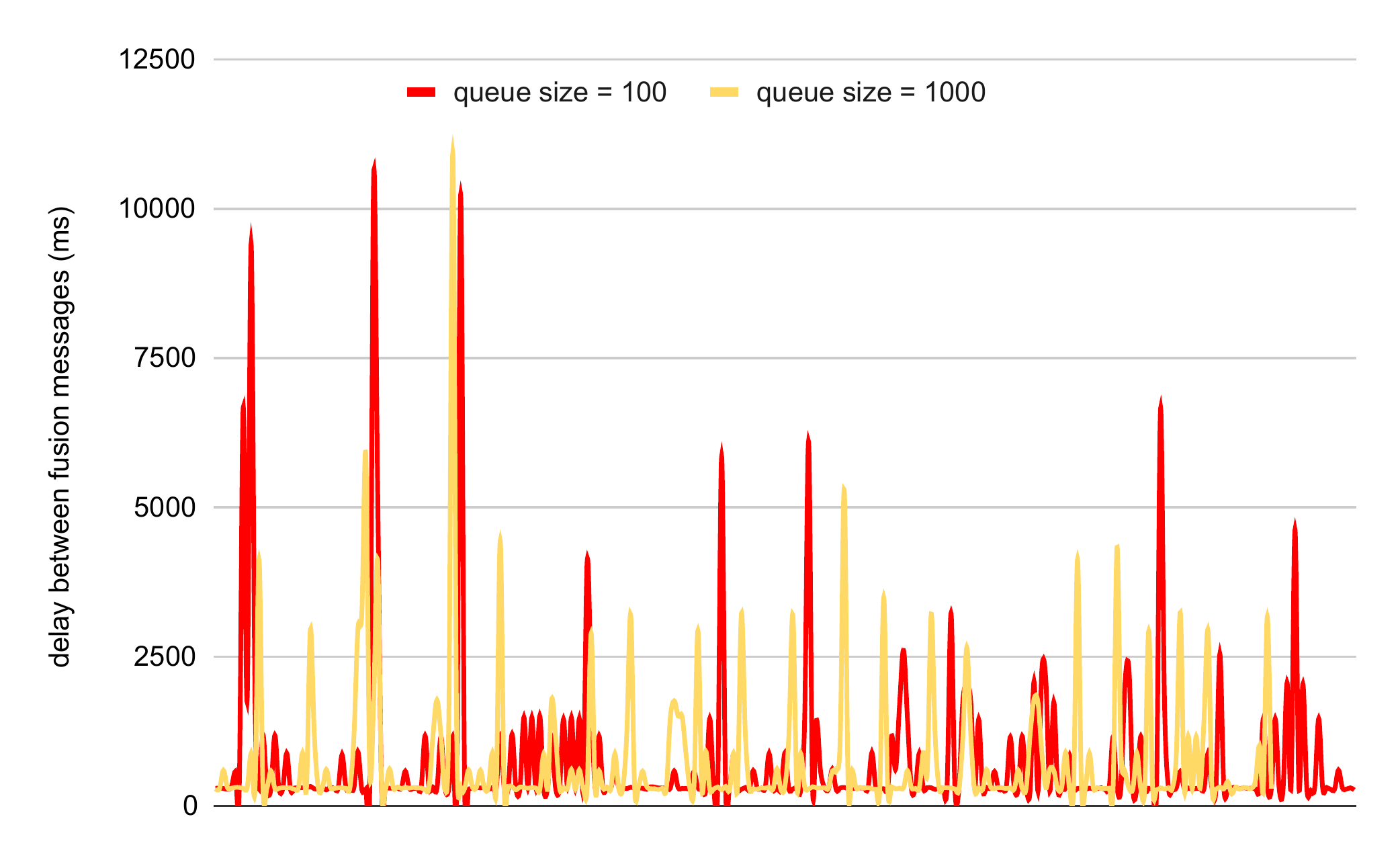}
	\caption{The delay between fusion messages.}
	\label{fig:ros-system-fusion}
\end{figure}

To understand the real impact of this enormous variation to the perception system, we record the timestamps for the \textit{/fusion} node to get synchronized messages from three perception modules and generate the fusion data. Figure~\ref{fig:ros-system-fusion} shows the delays between the fusion messages in milliseconds. Queue size with 100 and 1000 of the buffer within the synchronizer are tested, respectively. The delay variations under queue size with 1000 are less than that with 100. For a queue size of 100, we can found that there are tremendous variations for the delay. The worst case of the fusion message's delay goes to over 10000ms, which means that the control module gets the lane or objects' perception results after 10 seconds since the camera captures objects or lane. The average and minimum value of the delay also goes to 773ms and 242ms, respectively. This huge variation makes the control of the vehicle very challenging because it is hard to predict.

\textbf{Insight 6:}~\textit{Communication middleware like ROS brings huge extra variations to the end-to-end autonomous driving systems. Long-tail latency exists, and it accumulates with competing with concurrent tasks. Adding queue size inside the synchronizer helps to reduce the variations.}



\section{Related Work}
\label{related-work}

Autonomous vehicles are proposed to understand the environment and drive without human intervention~\cite{liu2019edge}. DNNs play an essential role in the sensing, perception, decision, and control tasks in autonomous driving. Generally, the research on DNNs for autonomous driving can be divided into two categories: training DNN models with higher accuracy and improving the runtime performance of trained DNN models.

Many DNN-based algorithms are deployed in autonomous vehicles for object detection, lane detection, semantic segmentation, localization, etc.~\cite{liu2020computing}. The object detection algorithms can be divided into two types: one-stage based algorithms like YOLO and SSD~\cite{redmon2018yolov3,liu2016ssd}; two-stage based algorithms like Fast R-CNN, Mask R-CNN, etc.~\cite{ren2015faster,he2017mask}. The key difference is whether there is a proposal bounding box stage. Semantic segmentation is used to detect driving segments. The fully convolutional neural network has been applied and achieves good performance~\cite{long2015fully}. LaneNet is a lane detection algorithm that uses an instance segmentation problem and applies image semantic segmentation algorithms~\cite{neven2018towards}. Another approach called PINet adds key points estimation with the instance segmentation~\cite{ko2020key}.

After the DNN models get trained, optimizing the model inference in latency, energy consumption, and memory utilization becomes a big challenge. In 2015, Han \textit{et al.} \cite{han2015learning} proposed pruning redundant connections and retraining the deep learning models to fine-tune the weights effectively, reducing computing complexity. Reducing the precision of operations and operands is another direction for the runtime optimization of DNN inference. Reducing precision is usually achieved by reducing the number of bits/levels representing the data, decreasing the computation requirements and storage costs~\cite{sze2017efficient}. Besides, the profiling of the DNN inference also gets more attention~\cite{wu2019machine}. MLPerf targets a uniformed profiling benchmark for the machine learning algorithms at the edge~\cite{reddi2020mlperf}.

\section{Conclusion}
\label{Conclusion}
DNNs are widely used in autonomous driving due to their high accuracy for perception, decision, etc. Understanding the variation of the DNN inference of autonomous driving becomes a fundamental challenge in real-time and efficient scheduling. Non-negligible time variations are observed in DNN inference, which significantly challenges scheduling safety-critical tasks. Therefore, in this work, we analyze the time variation in DNN inference in fine granularity and derive six insights into understanding DNN inference time variations.




\bibliographystyle{plain}
\bibliography{main}

\end{document}